\documentclass[conference, 10pt]{IEEEtran}
\usepackage{cite}
\usepackage{amsmath,amssymb,amsfonts}
\usepackage{algorithmic}
\usepackage{graphicx}
\usepackage{textcomp}
\usepackage{xcolor}
\usepackage{subcaption}
\usepackage{color,soul}
\usepackage{float}
\usepackage{romannum}
\usepackage{breqn}
\usepackage{hyperref}

\hypersetup{
    colorlinks=true, %set true if you want colored links
    linktoc=all,     %set to all if you want both sections and subsections linked
    linkcolor=red,  %choose some color if you want links to stand out
    citecolor=green
}

\usepackage{adjustbox}

\begin{document}

\title{Deep Joint Transmission-Recognition for Multi-View Cameras}

\author{
\IEEEauthorblockN{Ezgi \"Ozy{\i}lkan}
\IEEEauthorblockA{\textit{Imperial College London} \\
London, UK \\
ezgi.ozyilkan17@imperial.ac.uk}
\and
\IEEEauthorblockN{Mikolaj Jankowski}
\IEEEauthorblockA{\textit{Imperial College London} \\
London, UK \\
mikolaj.jankowski17@imperial.ac.uk}
}

\maketitle

\begin{abstract}
We propose joint transmission-recognition schemes for efficient inference at the wireless edge. Motivated by the surveillance applications with wireless cameras, we consider the person classification task over a wireless channel carried out by multi-view cameras operating as edge devices. We introduce deep neural network (DNN) based compression schemes which incorporate digital (separate) transmission and joint source-channel coding (JSCC) methods. We evaluate the proposed device-edge communication schemes under different channel SNRs, bandwidth and power constraints. We show that the JSCC schemes not only improve the end-to-end accuracy but also simplify the encoding process and provide graceful degradation with channel quality. 

\end{abstract}


\section{Introduction}
\label{sec:intro}
The number of Internet of Things (IoT) devices has been growing expeditiously across the globe. From smart home devices to wearable technologies, the applications of IoT devices are diverse as well as unprecedented. As of $2019$, the percentage of businesses that make use of IoT technologies has increased from $13\%$ in $2014$ to about $25\%$ \cite{mck}. As each generation of the IoT devices comes with more reliable connectivity, better sensors, more computing power and greater data storage capacity, the growth of IoT technologies is expected to accelerate even further. 

%Equipped with reliable connectivity, high-tech sensors, more computing power and greater data-storage capacity, the growth of IoT technologies is expected to accelerate even further. 

Currently, most IoT devices operate as wireless sensors at the edge: they collect data, preprocess it and offload to an edge (or cloud) server. One of the main challenges in such setting is that the IoT devices are typically power and memory constrained -- meaning they can only carry out a limited amount of computations, which makes the resource allocation problem challenging. In the scenario considered in this paper, the computations of interest are imposed by a deep neural network (DNN). In this work, we consider multi-view cameras, whose fields of vision overlap, to carry out person classification task at the wireless network edge\footnote{For a more formal construction, let us denote $\mathbf{x}_{1}$,$\mathbf{x}_{2}$ as a pair of images from two different cameras and $H(\mathbf{x}_{1},\mathbf{x}_{2})$ as their joint entropy. Given that the contents of these images are correlated due to overlapping fields of vision of the cameras, we have:
    \begin{equation*}
        I(\mathbf{x}_{1},\mathbf{x}_{2})=H(\mathbf{x}_{1})+H(\mathbf{x}_{2})-H(\mathbf{x}_{1},\mathbf{x}_{2}) > 0
    \end{equation*} Motivated by this observation, we seek a compression model to jointly compress the features of these correlated images.}. 
Additionally, since the cameras and the edge server communicate through a wireless channel, this transmission poses another challenge as it establishes a significant bottleneck as well as leads to latency and consumption of considerable energy by the resource-constrained cameras.

Some recent works \cite{edge:2}, \cite{edge:3}, \cite{mj:1} demonstrate the potential of splitting DNNs between the edge devices and the server. The practical objective of splitting the network is to make sure that the subnetwork deployed on IoT devices accommodates for their inherent limited computational and storage resources. 

Furthermore, although Shannon's \emph{separation theorem} proves the optimality of the separate design of source and channel codes in the infinite blocklength regime, the joint source-channel coding (JSCC) is known to improve the performance and robustness in practical communication systems \cite{jscc:1}. Considering real-time constraints, this makes deep JSCC scheme attractive for realistic distributed learning and inference scenarios such as the one considered in this work. Our contributions can be summarized as follows:
\begin{itemize}
    \item We propose a DNN training strategy for joint device-edge classification task using deep JSCC.
    \item We propose an autoencoder-based architecture for intermediate feature compression under noisy and bandwidth-limited channel conditions.
    \item We evaluate the proposed schemes on person classification task, under different channel SNRs, bandwidth and power constraints. 
\end{itemize}

\section{Methods}

\begin{figure*}[t]
        \centering
        \includegraphics[width=0.89\textwidth]{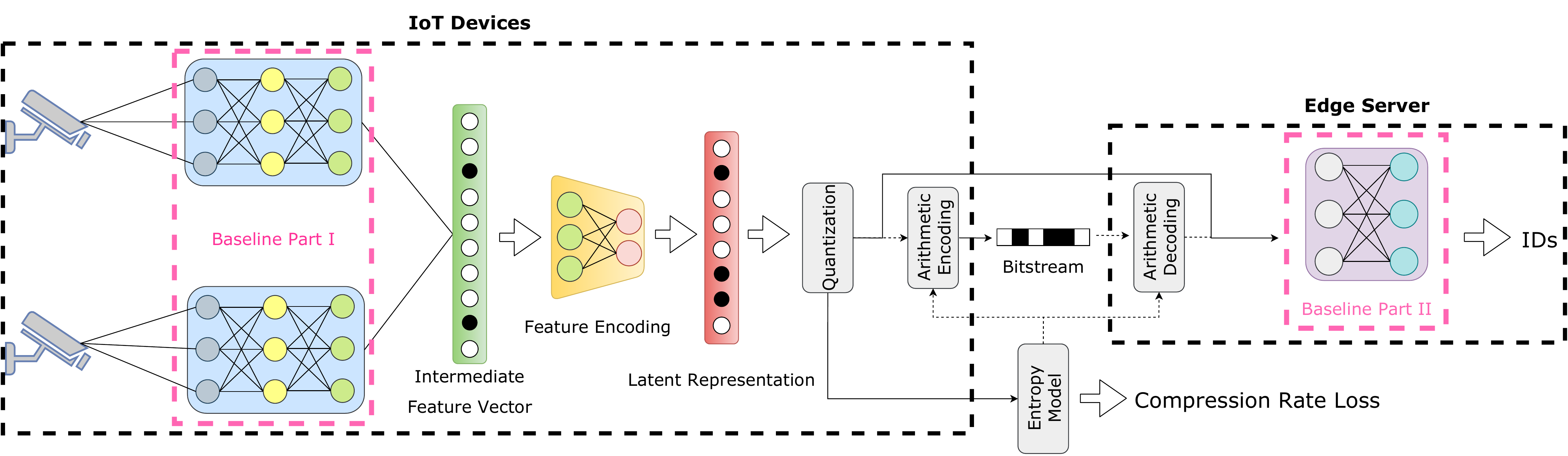}
\caption{An overview of the digital (separate) transmission scheme. The intermediate feature vector is compressed using a single fully-connected layer. At the server side, the quantized latent representation is used in order to predict the unique person IDs. Arithmetic coding is only carried out during testing, as indicated with dashed lines in the figure. For the sake of obtaining an upper bound on the performance, ideal channel coding scheme is assumed.}
\label{fig:digital}
\end{figure*}

In this section, we analyze two different paradigms for the task of person classification at the wireless edge carried out by multi-view cameras:
digital (separate) scheme, shown in Fig. \ref{fig:digital}, and JSCC-based approaches, shown in Fig. \ref{fig:sep}, \ref{fig:jdec} and \ref{fig:systems}. Section \ref{sec:classification_baseline} describes the classification baseline, which is used both by the digital and JSCC schemes that are described in Section \ref{sec:digital} and \ref{sub:jscc}, respectively. The section concludes with a description of multi-step training strategy, which was shown to be well-suited for the JSCC approaches.

Considering the nature of surveillance applications, stringent bandwidth values will be considered -- this means that features output by a DNN, instead of the original images, will be transmitted over the wireless channel. Since extracting such feature vectors increases the computational load on edge devices, we aim to balance the trade-off between on-device computations and communication overhead. Although any differentiable channel model can be employed to train the JSCC approaches, we consider an additive white Gaussian noise (AWGN) channel in this paper. Formally, let $\mathbf{x} \in \mathbb{R}^{B}$ and $\mathbf{y} \in \mathbb{R}^{B}$ be the channel input and channel output vectors, respectively. The channel output is then calculated as $\mathbf{y}=\mathbf{x}+\mathbf{z}$, where $\mathbf{z} \in  \mathbb{R}^{B}$ and $z_{i} \sim \mathcal{N}(0,\sigma_{noise}^{2})$. For every channel input vector, we impose an average power constraint of $P=1$, i.e. $\frac{1}{B} \sum _{i=1} ^{B} x_{i}^{2} \leq P$. Accordingly, the channel SNR is then defined as:

\begin{equation}
    \mathrm{SNR}=10\log_{10} \left(\frac{P}{\sigma_{noise}^{2}}  \right) \hspace{0.2cm}    \mathrm{(dB). }
\label{eq:snr}
\end{equation}
To compare JSCC approaches with the digital (separate) alternative, Shannon capacity formula is used, that is:
\begin{equation}
    C=\frac{1}{2}\log_{2}\left(1+\frac{P}{\sigma_{noise}^{2}}\right),
\label{eq:shannon}
\end{equation}
%Equation \eqref{eq:shannon} yields us the minimum $SNR$ requirement that allows us to send $C$ bits at the \hl{channel usage of $B$ times Change??}, where $B$ corresponds the to available channel bandwidth of real values. 
which yields the minimum SNR requirement that allows us to send $C$ bits through the channel, under the assumption that the ideal source and channel codes are in use.

\subsection{Classification baseline}
\label{sec:classification_baseline}

\begin{figure}[H]
        \centering
        \includegraphics[width=0.48\textwidth]{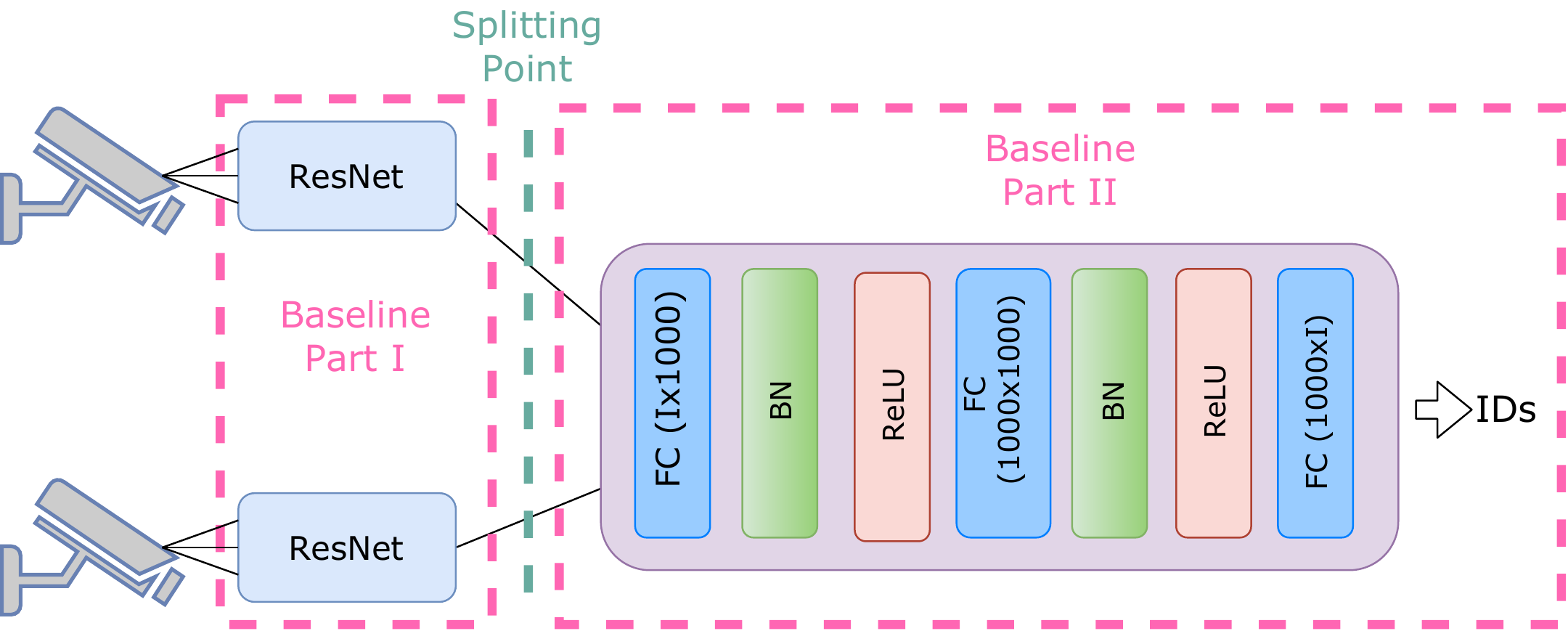}
    \caption{Proposed baseline architecture for classification. Fully-connected layer parameters are denoted as: input size $\times$ output size. For two-camera setting, I in the figure corresponds to $\mathrm{I} = \mathrm{I_{1}} + \mathrm{I_{2}}$, where $\mathrm{I_{\mathcal{K}}}$ is equal to the number of the unique person IDs at Camera $\mathcal{K}$}.
    \label{fig:classif}
\end{figure}

\begin{figure*}[t]
        \centering
        \includegraphics[width=0.8\textwidth]{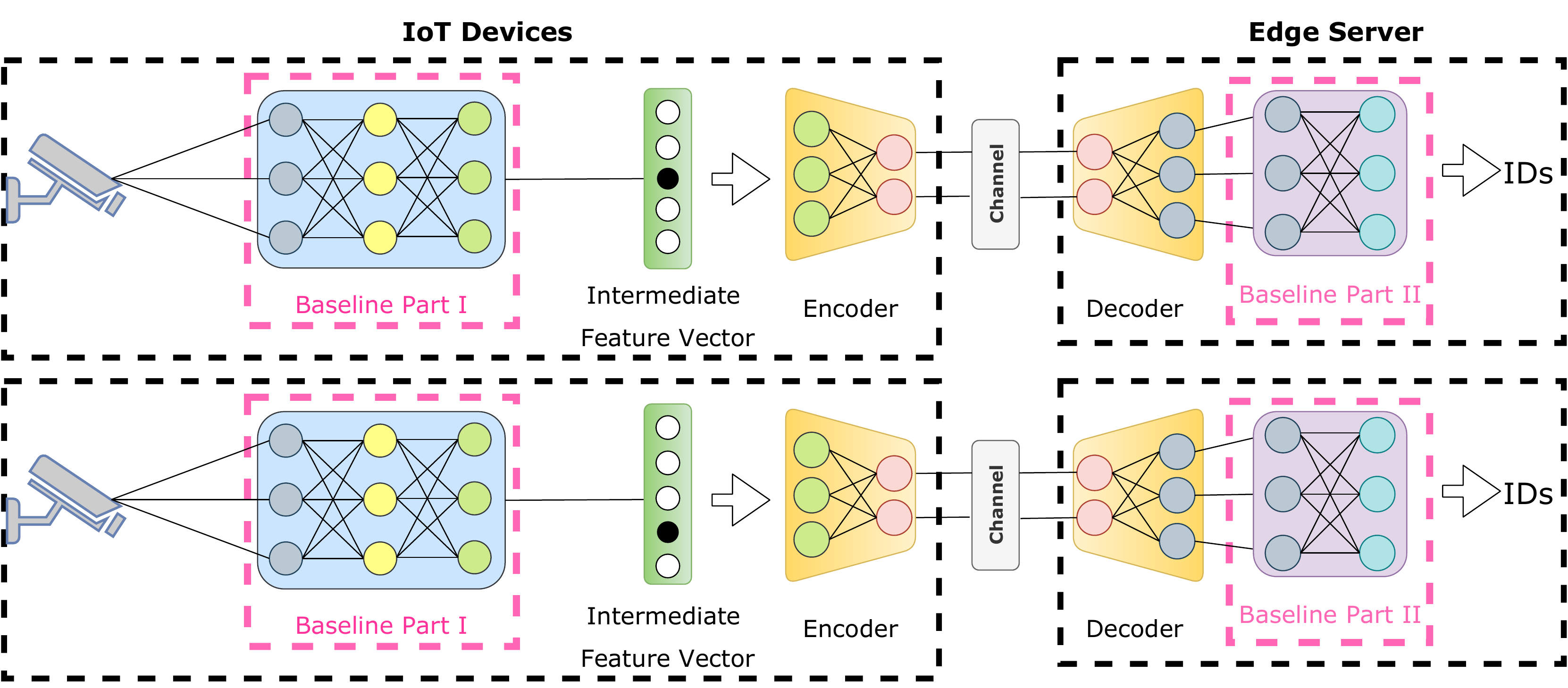}
\caption{An overview of the single user schemes for both cameras.  Similar to the models in Fig. \ref{fig:jdec} and \ref{fig:systems}, the baseline is split between the IoT devices and the edge server at the same splitting point indicated in Fig. \ref{fig:classif}. The intermediate feature vector  is compressed by the encoder shown in Fig. \ref{fig:ae} and sent through a wireless channel. Note that in the figure, the autoencoders  are independent of each other and the channels incorporated are orthogonal with respect to one another. Furthermore, an average power constraint of  $P=1$ is imposed for each user. The total channel bandwidth for the single user scheme shown in the figure is calculated as $B=B_{1}+B_{2}$, where $B_{\mathcal{K}}$ is the bandwidth (the maximum number of channel uses per transmission) allocated for the Camera $\mathcal{K}$. We will refer to this model as the \emph{Single Users} in the following parts.}
\label{fig:sep}
\end{figure*}

\begin{figure*}[t]
        \centering
        \includegraphics[width=0.8\textwidth]{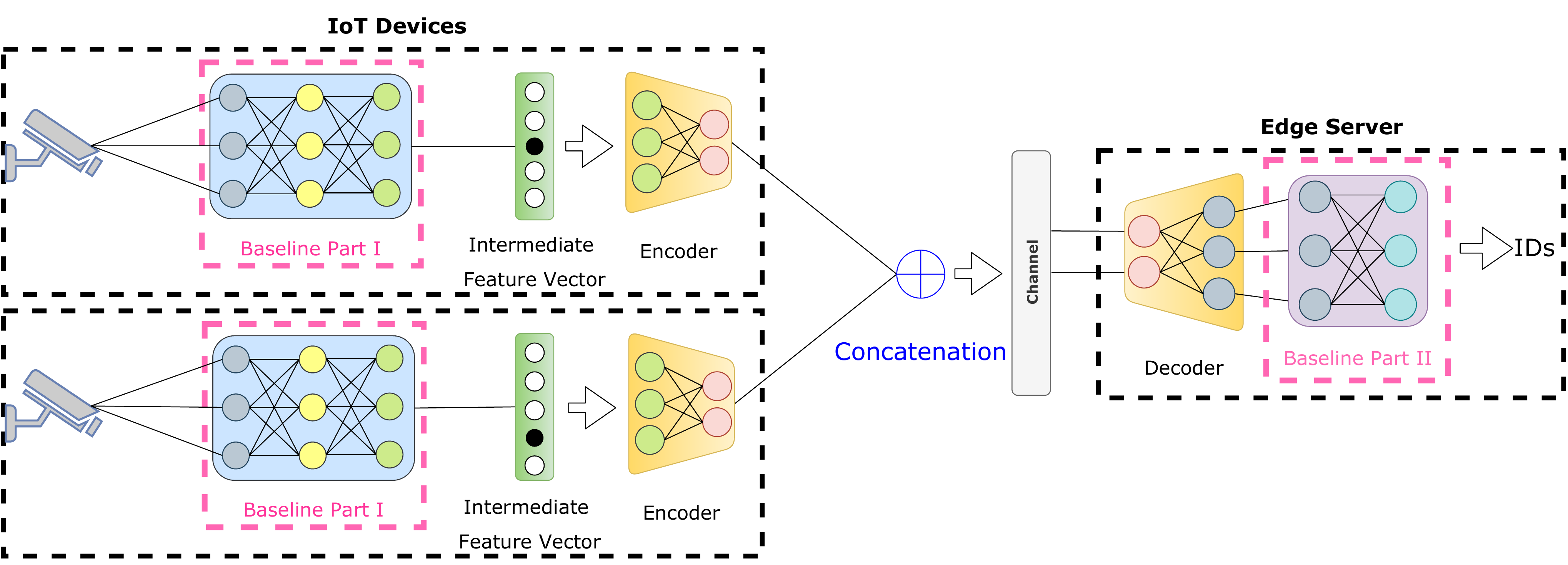}
\caption{An overview of the evaluated joint decoding schemes. The baseline is split between the IoT devices and the edge server at the same splitting point indicated in Fig. \ref{fig:classif}. The intermediate feature vector  is compressed and sent through the wireless channel by the autoencoder architecture in Fig. \ref{fig:ae}. Note that for the \emph{Joint Decoding + Orthogonal Multiple Access} scheme, the bandwidth allocation is set to be equal for both users. For the implementation of the \emph{Joint Decoding + Non-Orthogonal Multiple Access} scheme, we replace the concatenation operation in the figure by the element-wise summation of the output of two encoders, which both have the dimensionality of \emph{B} given a bandwidth budget of \emph{B}. Note that for the evaluated orthogonal and non-orthogonal multiple access approaches, we impose an average power constraint of $P=1$ for every channel input vector and also an equal power allocation for both users. In the following parts, we will refer to these models as the \emph{J-Dec + OMA} and \emph{J-Dec + NOMA} for the cases of orthogonal and non-orthogonal multiple access schemes, respectively.}
\label{fig:jdec}
\end{figure*}

\begin{figure*}[t]
    \begin{subfigure}[b]{\linewidth}
        \centering
        \includegraphics[width=0.7\textwidth]{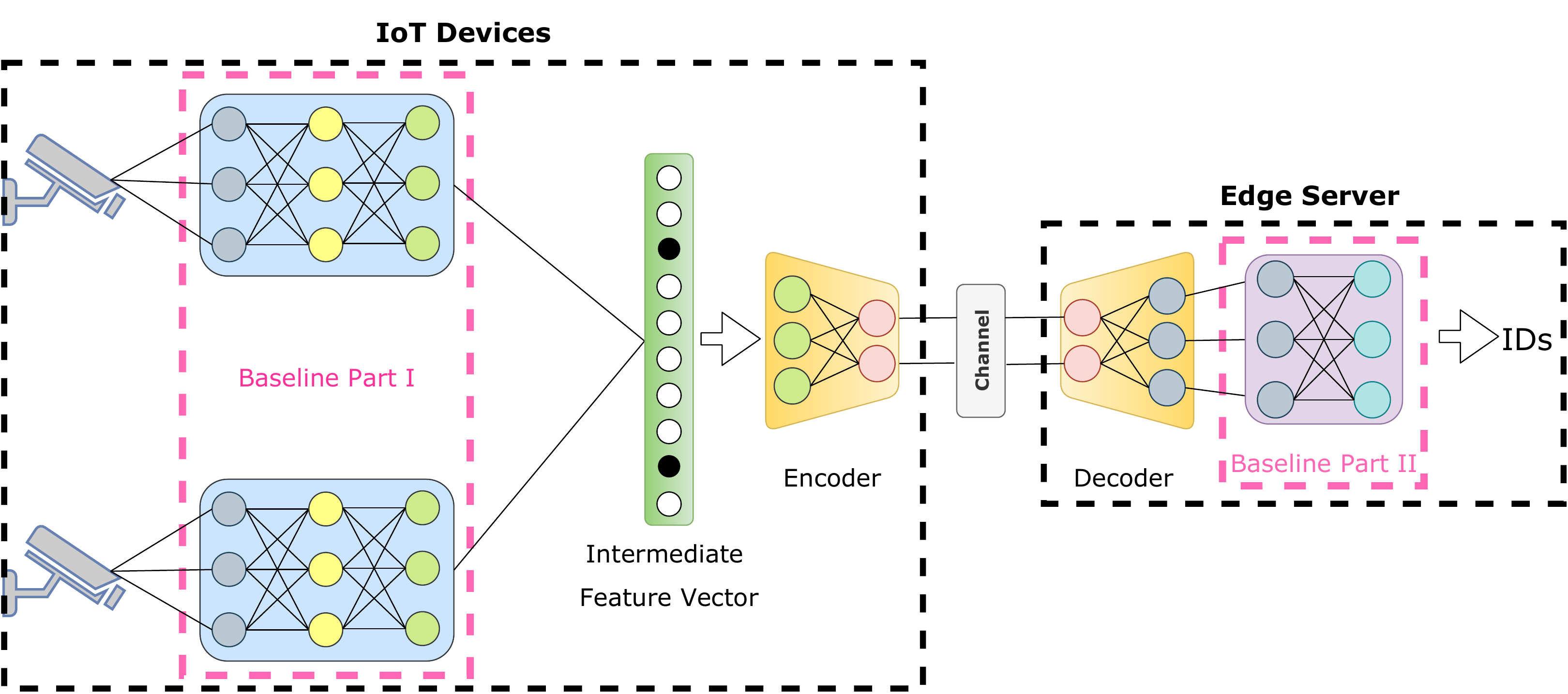}
        \caption{\emph{Joint Encoding}}
        \label{fig:joint}
    \end{subfigure}
    \hspace{1pt}
    \begin{subfigure}[b]{\linewidth}
        \centering
        \includegraphics[width=0.89\textwidth]{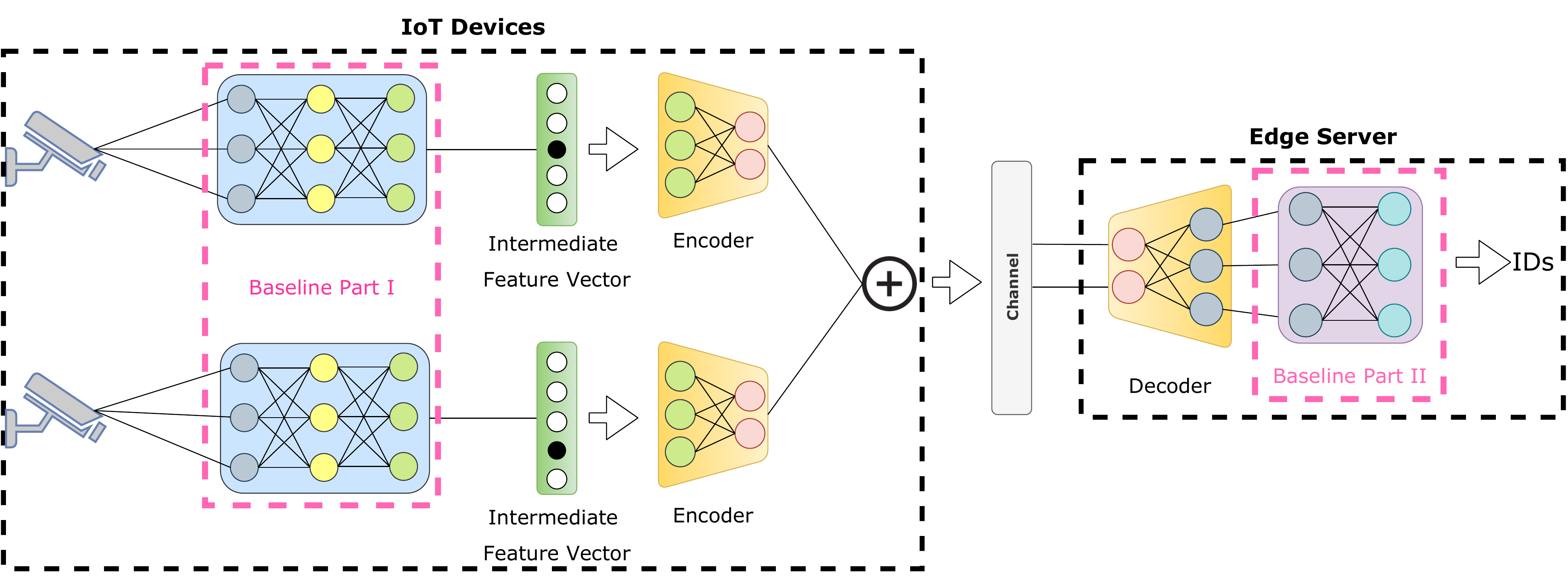}
        \caption{\emph{Joint Encoding + Non-Orthogonal Transmission}}
        \label{fig:nonorth}
    \end{subfigure}
\caption{An overview of the evaluated joint encoding schemes. The baseline is split between the IoT devices and the edge server at the same splitting point for both models indicated in Fig. \ref{fig:classif}. The intermediate feature vector  is compressed and sent through the wireless channel by the autoencoder architecture in Fig. \ref{fig:ae}. At the server side, the received feature vector is decoded and the edge server completes the forward pass in order to predict the unique person IDs. Unlike the \emph{Joint Encoding}  scheme, the \emph{Joint Encoding + Non-Orthogonal Transmission} model employs two distinct encoders. The channel symbols for Fig. \ref{fig:nonorth} is implemented as $\mathbf{s}=\mathbf{s_{1}}+\mathbf{s_{2}}$, where $\mathbf{s_{\mathcal{K}}}$ denotes the channel symbols for Camera $\mathcal{K}$. We will refer to these models shown in Fig. \ref{fig:joint} and \ref{fig:nonorth} as the \emph{J-Enc1} and \emph{J-Enc2} in the following parts, respectively.}
\label{fig:systems}
\end{figure*}

Given an image, our goal is to identify the people appearing in that image. For this task, we employ a DNN that outputs a multi-hot encoded vector. For the classification baseline, we employ a pretrained ResNet-18 \cite{resnet} for each camera followed by three fully-connected layers with  Batch Normalization (BN) and rectified linear unit (ReLU) activation layers added after each of them (see Fig.  \ref{fig:classif}). We will refer to this architecture as the \emph{classification baseline} in the following parts.

%Note that for the evaluated orthogonal and non-orthogonal multiple access approaches, we impose an average power constraint of $P=1$ for every channel input vector and either impose an equal power allocation or let the network learn how to do the power allocation for both users. In the following parts, we will refer to these model as the \hl{\emph{J-Dec + OMA1}} and \hl{\emph{J-Dec + NOMA1}} for the case of equal power allocation, and as the \hl{\emph{J-Dec + OMA2}} and \hl{\emph{J-Dec + NOMA2}} for the case of learned power allocation

\subsection{Digital (Separate) Transmission of Extracted Features}
\label{sec:digital}

An overview of the digital (separate) scheme is provided in Fig. \ref{fig:digital}. After using the Baseline Part \Romannum{1} to extract the intermediate feature vector, we pass it through the lossy feature encoder, which consists of a single fully-connected layer. The obtained latent representation is then quantized, compressed by the arithmetic coder and transmitted through the channel to the edge server, where it is passed through the Baseline Part \Romannum{2} in order to predict the unique person IDs.

To ensure that the quantization step is differentiable, we adopt the quantization noise from \cite{quant} during the training phase. Specifically, we add a uniform noise to each element in the latent representation, as follows:
\begin{equation}
    Q(\mathbf{r}) = \mathbf{r} + \mathcal{U}\left(-\frac{1}{2}, \frac{1}{2}\right), 
\end{equation}
where $Q(\cdot)$ is the approximated quantization, $\mathbf{r}$ is the latent representation and $\mathcal{U}(\cdot, \cdot)$ is the uniform noise vector. During inference, each element in the latent representation rounded to the nearest integer on which the arithmetic coding is subsequently performed.

The arithmetic coding we implement is based on estimating the distribution of the quantized latents. Assuming that the vector elements $q_{i} \in \mathbf{q}=Q(\mathbf{r})$ are i.i.d with some probability mass function (PMF) of $p(q)$, we first-order approximate $p(q)$ as a continuous-valued probability density function $p_c(q)$ as:

\begin{equation}
    p_c(q) = \sum_{k=1}^K \alpha_k \frac{1}{\sigma_k \sqrt{2\pi{}}} e^{-\frac{1}{2}\left(\frac{q-\mu_k}{\sigma_k}\right)^2},
\end{equation}
where $K$ is the number of Gaussian mixtures, $\sigma_k$ are mixture scales, $\mu_k$ are mean values and $\alpha_k$ are the corresponding mixture weights. We experiment with $K \in \{1, 2, 3\}$ in this paper since these values are found to be performing well for small $C$ values in Equation \eqref{eq:shannon}. Finally, we evaluate PMF $p(q)$ at discrete values $q \in \mathbb{Z}$ by integrating $p_c(q)$ over $\left[q-\frac{1}{2}, q+\frac{1}{2}\right]$ in order to obtain:

\begin{equation}
p(q) = \int_{q - \frac{1}{2}}^{q + \frac{1}{2}}p_c(x) dx = 
F_c\left(q + \frac{1}{2}\right) - F_c\left(q - \frac{1}{2}\right),
\end{equation}
where $F_c$ is the cumulative density function of the distribution $p_c(q)$.

%from \cite{jscc:3}, which is based on a prior probability model on the quantized latent representation that is learned by the entropy model. \hl{we are no more using Balle actually, can we keep it though?} Probability distribution estimated as such is used to compute Shannon entropy of the latent representation. We then use this entropy to determine the compression rate loss, which is based on the number of bits required per picture calculated using the entropy model. 

The arithmetic encoded version of the quantized latent representation is then transmitted to the server using a channel code. Since any channel coding scheme introduces some degree of error, we assume capacity achieving channel codes over the channel in order to obtain an upper bound on the performance for the digital scheme that employs the architecture shown in Fig. \ref{fig:digital}.

\subsubsection{Loss Function}

In order to achieve a reasonable performance trade-off between the person classification task and the compression rate, we define the loss as weighted sum of the two objectives, which we aim to minimize:
\begin{equation}
    L = l_{\mathrm{cross-entropy}}  - \lambda \cdot \log_{2}p(\mathbf{q}),
    \label{eq:weighted}
\end{equation}
where $l_{\mathrm{cross-entropy}}$ and $p(\mathbf{q})$ refer to cross-entropy loss between predicted IDs and ground truth for person classification task, and the PMF of the quantized vector $\mathbf{q}$ respectively. In this work, we experiment with small $\lambda$ values such that $\lambda \in [0.001, 0.045]$ in order to obtain a reasonable accuracy. We also allow different dimensionality for the latent representation, between 4 and 16.

\subsubsection{Training Strategy}

The entire network in Fig. \ref{fig:digital} is trained end-to-end with the cross-entropy loss for $50$ epochs, using SGD with Nesterov momentum of $0.9$, learning rate of $0.01$ and $\mathit{L}_{2}$ penalty weighted by $5 \cdot 10^{-4}$. We reduce the learning rate by a factor of $0.1$ every $10$ epoch. Note that the arithmetic coding is implemented only for testing and therefore bypassed during the training phase.

%Our framework is flexible, and can be easily adapted to any system that incorporates DNNs. We focus on image classification task as it is the most frequent approach to automatically analyse image content and generate its metadata. Given an image and a finite set of possible classes, the classification task aims at assigning the correct class label to the image. We experiment with VGG16 \cite{vgg} with batch normalization (BN) added after each convolutional layer as it is one of the most popular networks employed for image classification. The network consists of 13 convolutional layers with stride 1 divided into 5 blocks, where each block is followed by a pooling operation. We consider each of the pooling operations as a potential network splitting point as it provides feature compression by construction, and does not affect the accuracy. After the last pooling layer we also employ a fully-connected classifier consisting of three fully-connected layers, where the first two have the output size of 512 and the last one maps 512-dimensional vector to 100-dimensional class predictions.

%Self-referencing in Fig. captions is OK???
\subsection{JSCC of Extracted Features}
\label{sub:jscc}

Inspired by \cite{mj:1} and \cite{mj:2}, we evaluate four classification schemes at the wireless edge, which all include an autoencoder-based network for intermediate feature compression combined with deep JSCC scheme: device-edge models employing \emph{Joint Decoding + Orthogonal Multiple Access}, \emph{Joint Decoding + Non-Orthogonal Multiple Access}, \emph{Joint Encoding} and \emph{Joint Encoding + Non-Orthogonal Transmission}\footnote{Note that the use of the word \emph{joint} in this case does not refer to the JSCC scheme. From now on, we will use this word to refer to both concepts and we will not be explicit as long as the context dictates its meaning.}, shown in Fig. \ref{fig:jdec} and \ref{fig:systems}. By applying edge detection schemes onto multi-nodes of the network, we aim to the reduce channel bandwidth requirements, compared to a single node setup.

In order to make a fair comparison, we also investigate sending the feature vectors
separately. Referring to this as \emph{Single Users} device-edge model, an overview of the architecture is provided in Fig. \ref{fig:sep}.

\subsubsection{Autoencoder architecture}

\begin{figure}
    \centering
    \begin{subfigure}[b]{0.4\linewidth}
        \centering
        \includegraphics[width=\textwidth]{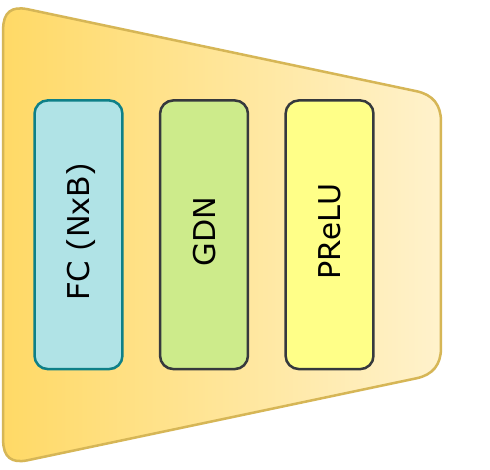}
        \caption{Encoder}
        \label{fig:encoder}
    \end{subfigure}
    \hspace{1pt}
    \begin{subfigure}[b]{0.55\linewidth}
        \centering
        \includegraphics[width=\textwidth]{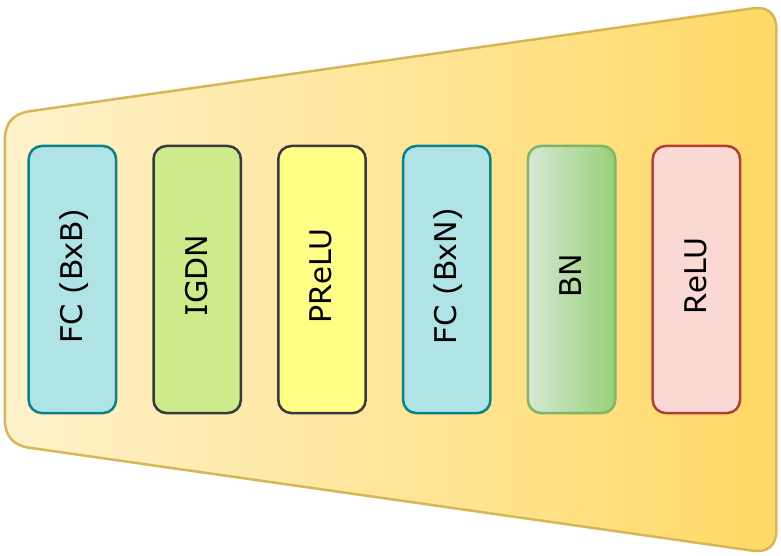}
        \caption{Decoder}
        \label{fig:decoder}
    \end{subfigure}
    \caption{Proposed autoencoder architecture for the JSCC schemes. Fully-connected layer parameters are denoted as: input size $\times$ output size. N, B and PReLU in the figure correspond to input dimension, channel bandwidth and parametric rectified linear unit, respectively.}
    \label{fig:ae}
\end{figure}

The proposed autoencoder architecture is shown in Fig. \ref{fig:ae}. Its goal is to compress the feature vectors in order to reduce communication overhead as well as the on-device computational load, thanks to its asymmetrical structure (similar to the one discussed in \cite{mj:1}). The fully-connected layers are either followed by Batch Normalization (BN), Generalized Divisive Normalization (GDN) or Inverse Generalized Divisive Normalization (IGDN) \cite{jscc:2, jscc:4} for the sake of Gaussianizing the data. We particularly employ GDN/IGDN layers since they are shown to be suitable for density modelling \cite{jscc:2, jscc:4} and are also used in the state-of-the-art image compression schemes such as \cite{jscc:3}.

\subsubsection{Training strategy}
\label{subs:training_strategy}

The training strategy for the evaluated JSCC approaches in Fig. \ref{fig:sep}, \ref{fig:jdec} and \ref{fig:systems} consists of three steps. Firstly, we pretrain the classification baseline in Fig. \ref{fig:classif} with cross-entropy loss for $30$ epochs, using SGD with Nesterov momentum of $0.9$, learning rate of $0.01$ and $\mathit{L}_{2}$ penalty weighted by $5 \cdot 10^{-4}$. The learning rate is reduced by a factor of $0.1$ every $10$ epoch. \par
Secondly, each image from the training set is passed through the Baseline Part \Romannum{1}, shown in Fig. \ref{fig:classif}, in order to extract all the possible intermediate feature vectors at the splitting point. These feature vectors are then used to pretrain the autoencoder, shown in Fig. \ref{fig:ae}, with $\mathit{L}_{1}$ loss for $50$ epochs, using SGD with Nesterov momentum of $0.9$, learning rate of $0.1$ and $\mathit{L}_{2}$ penalty weighted by $5 \cdot 10^{-4}$. The learning rate is reduced by a factor of $0.1$ after $20^{th}$ and $40^{th}$ epochs. Note that AWGN channel model is incorporated during training of the autoencoder so that the autoencoder is able to learn the robust transmission of feature vectors.

Thirdly, the entire network is trained end-to-end by combining Baseline Parts \Romannum{1} and \Romannum{2}, and placing the pretained autoencoder at the splitting point shown in Fig. \ref{fig:classif}. Similar to the first step,
the network is trained with cross-entropy loss for $50$ epochs, using SGD with Nesterov momentum of $0.9$, learning rate of $0.01$ and $\mathit{L}_{2}$ penalty weighted by $5 \cdot 10^{-4}$. Likewise, the learning rate is reduced by a factor of $0.1$ every $10$ epoch. 

Instead of single-step training the networks directly, the proposed multi-step training strategy allows the JSCC models to achieve superior classification accuracy, even with significantly low channel bandwidths (similar to the one discussed in \cite{jankowski2020wireless}).

\section{Results}

In this section, we discuss the performance of the evaluated JSCC approaches and compare it with the digital (separate) transmission paradigm. Before presenting the results, the experimental setup, along with the motives for the dataset choice, is discussed first.

\subsection{Experimental setup}

To assess the performance of the proposed architectures for the person classification task, we use the `WILDTRACK' dataset \cite{wild}, where there are $7$ static cameras whose fields of view are overlapping within one another\footnote{Annotated dataset can be downloaded from: \url{https://www.epfl.ch/labs/cvlab/data/data-wildtrack/}}. One of the main characteristics of the `WILDTRACK' dataset is that the cameras captured a realistic setup of walking pedestrians in front of the main building of ETH Zurich, Switzerland \cite{wild}. Furthermore, the high precision joint-camera calibration and synchronization of the `WILDTRACK` dataset surpasses those of the PETS $2009$ \cite{PETS}, which has been recognized as a challenging benchmark dataset at the time of publishing. Although the EPFL-RLC \cite{epfl}, another multi-view dataset, improves the joint-calibration accuracy and synchronization of multi-camera setup, compared to the PETS $2009$, it provides annotations only for a small subset of the total frames, making it unsuitable for deep-learning-based multi-view detection schemes \cite{wild}.

\begin{figure*}[tb]
\centering

\begin{subfigure}[t]{0.24\textwidth}
        \centering
    \includegraphics[width=\textwidth]{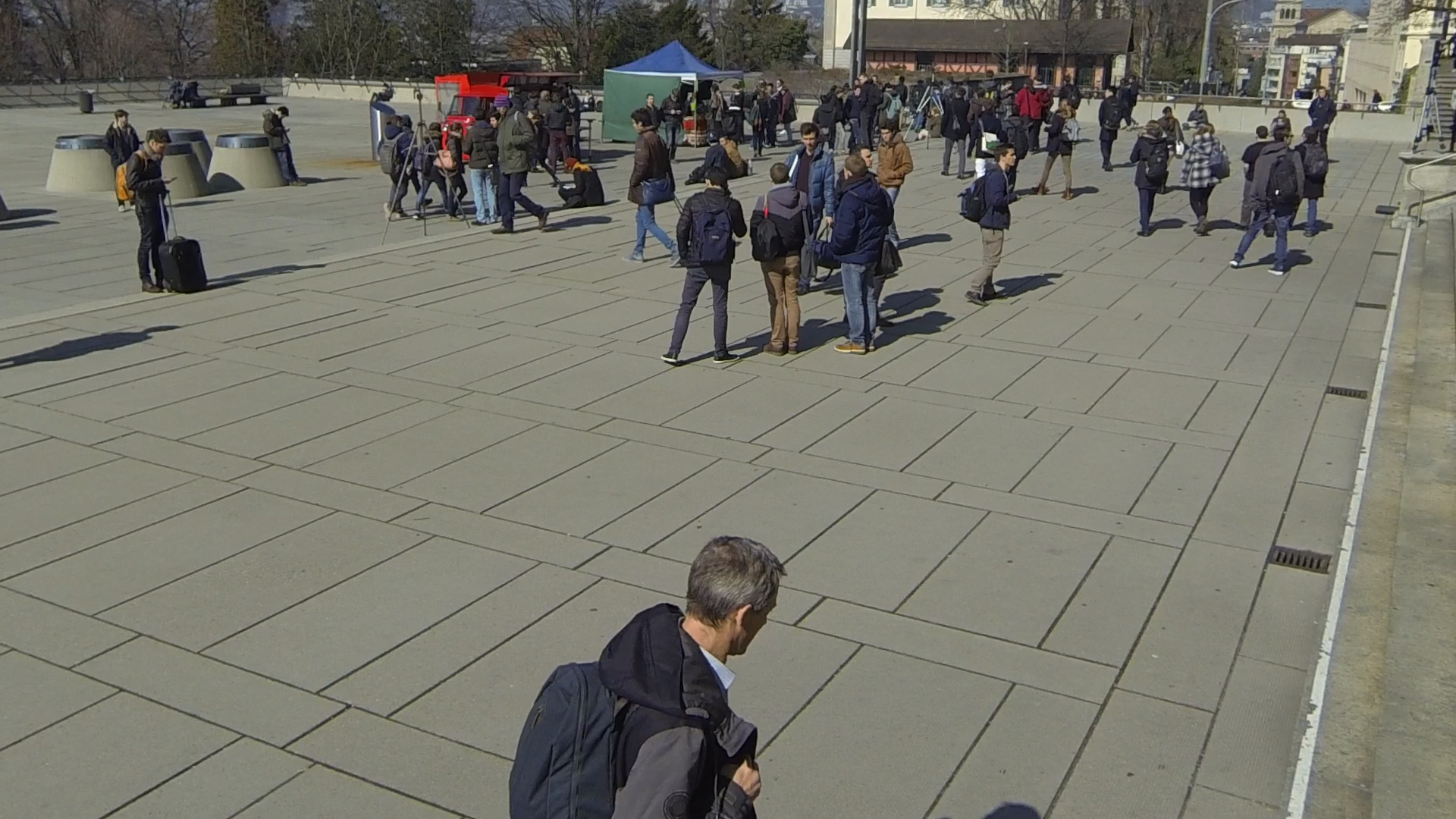}
    \caption{Camera 1.}
\end{subfigure}
    \hspace{.5pt}
\begin{subfigure}[t]{0.24\textwidth}
        \centering
    \includegraphics[width=\textwidth]{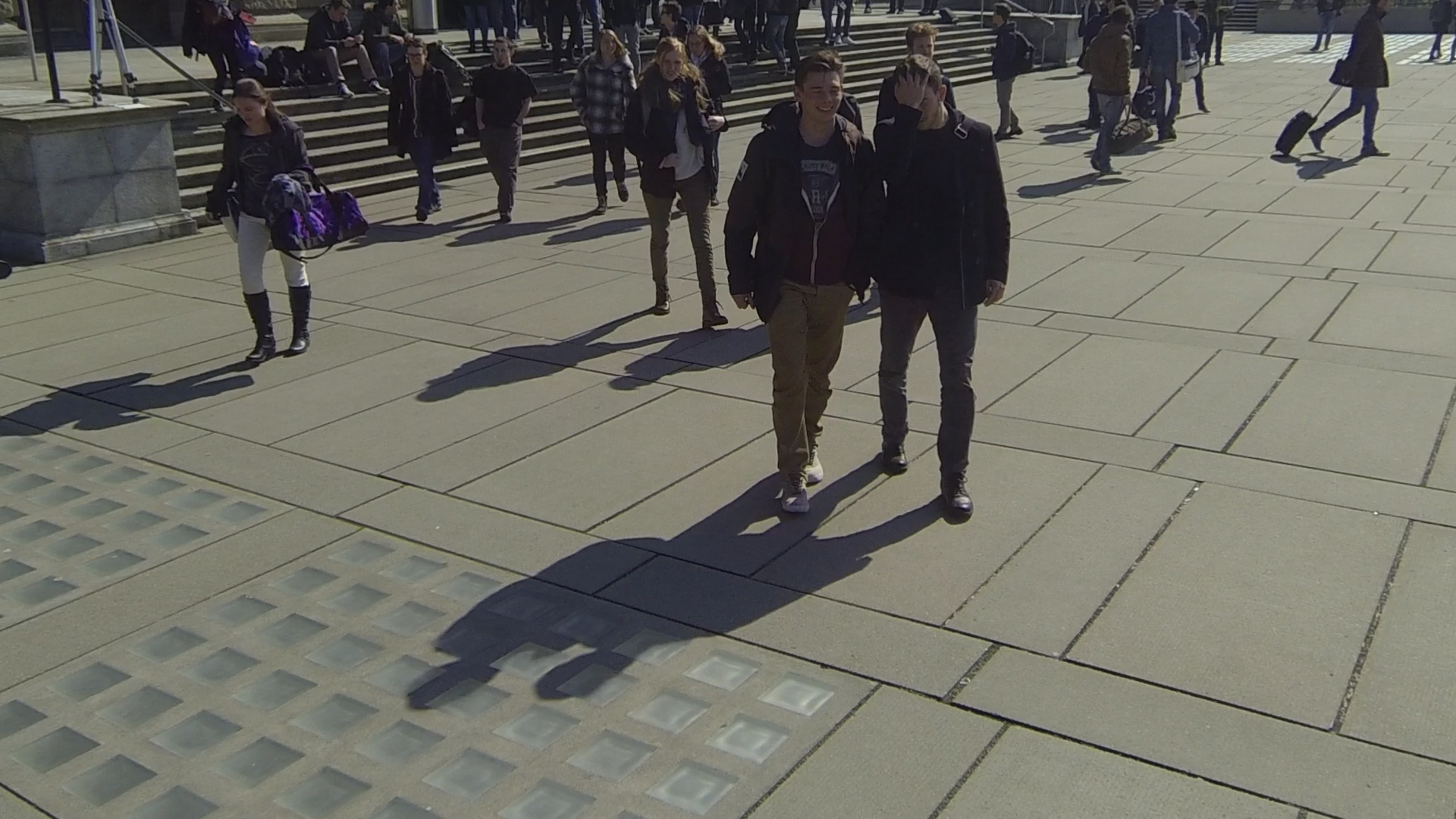}
    \caption{Camera 2.}
\end{subfigure}
    \hspace{.5pt}
\begin{subfigure}[t]{0.24\textwidth}
        \centering
    \includegraphics[width=\textwidth]{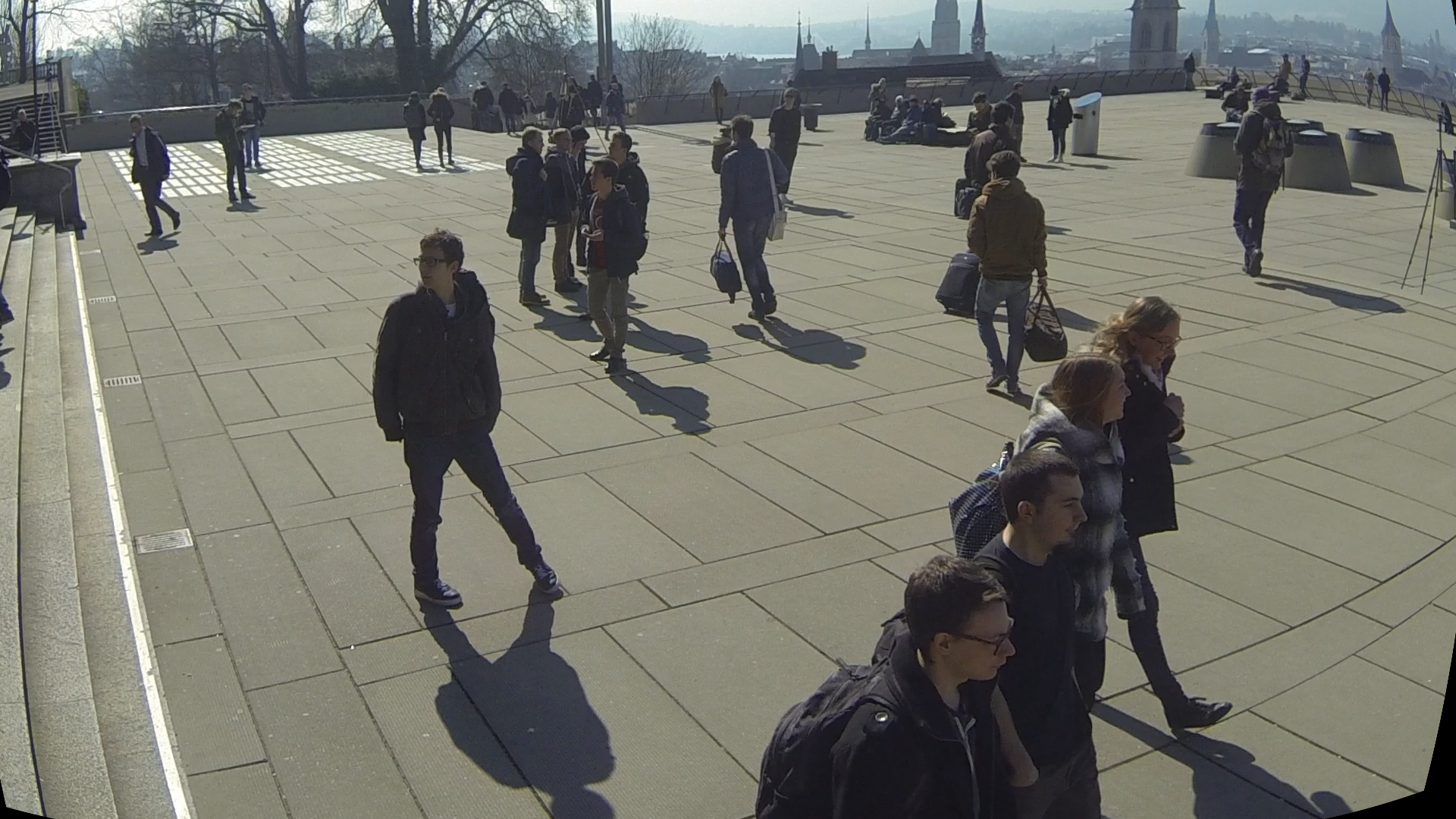}
    \caption{Camera 3.}
\end{subfigure}
    \hspace{.5pt}
\begin{subfigure}[t]{0.24\textwidth}
        \centering
    \includegraphics[width=\textwidth]{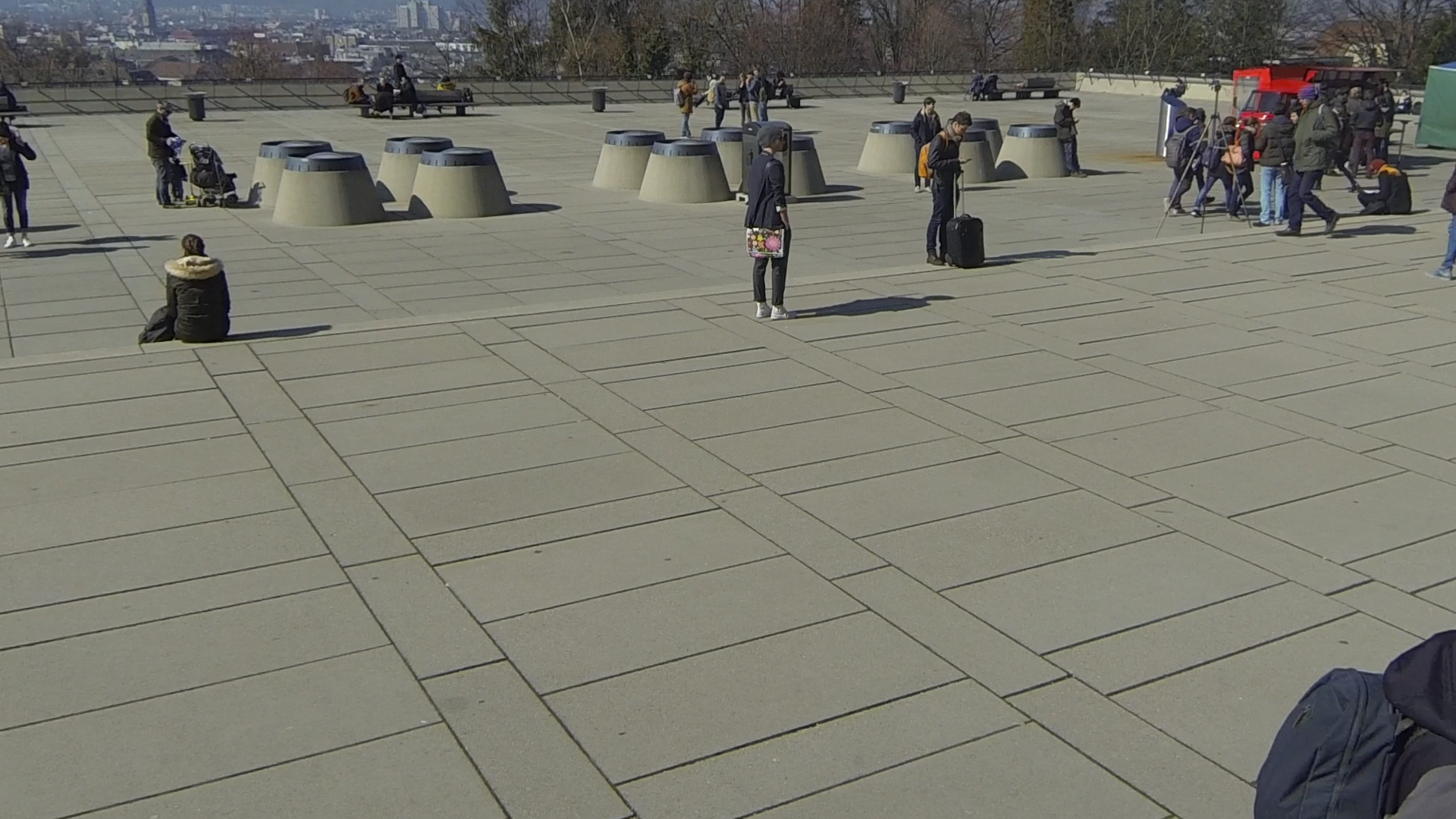}
    \caption{Camera 4.}
\end{subfigure}
    \hspace{1pt}
\begin{subfigure}[t]{0.24\textwidth}
        \centering
    \includegraphics[width=\textwidth]{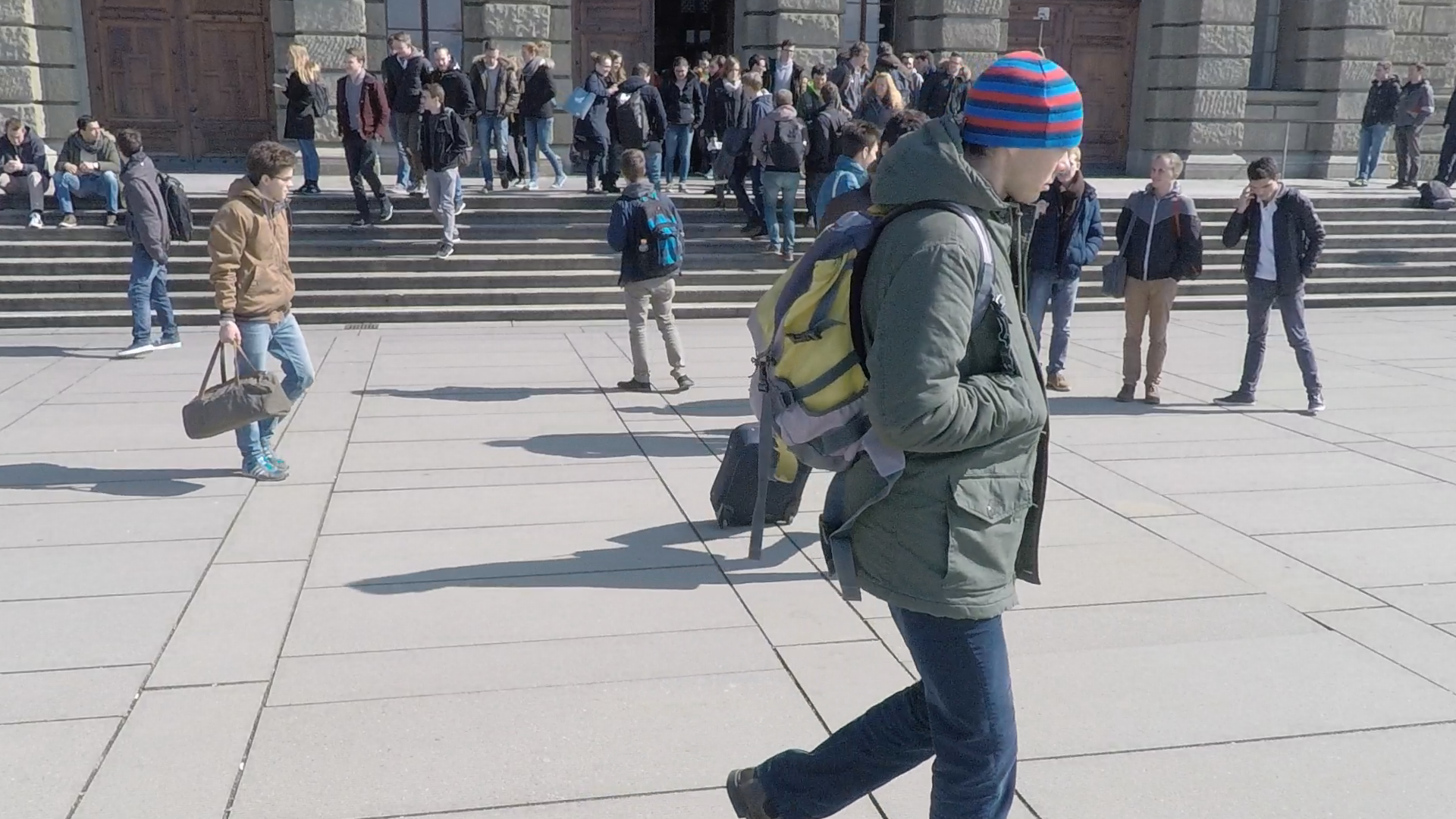}
    \caption{Camera 5.}
\end{subfigure}
    \hspace{1pt}
\begin{subfigure}[t]{0.24\textwidth}
        \centering
    \includegraphics[width=\textwidth]{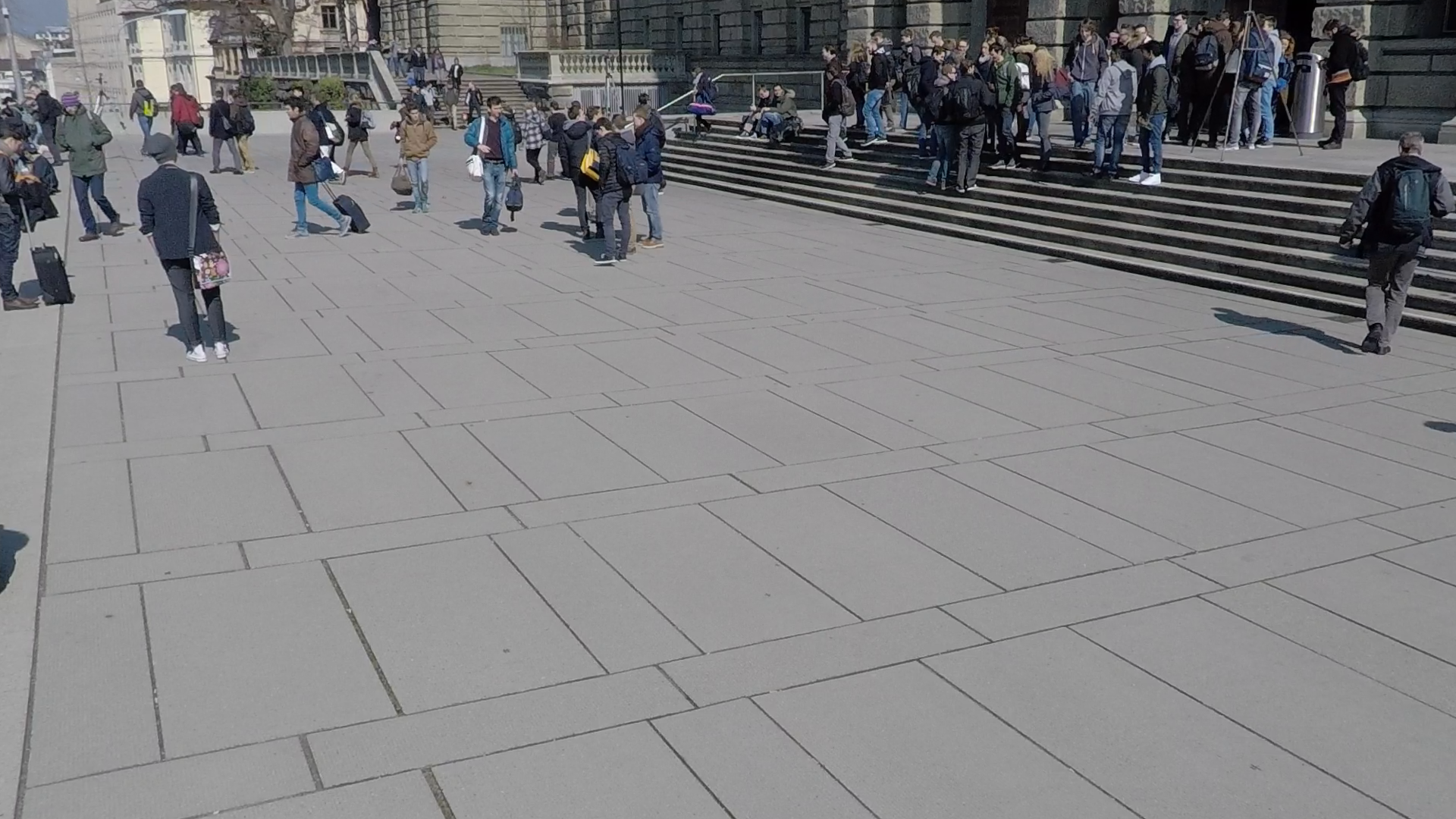}
    \caption{Camera 6.}
\end{subfigure}
    \hspace{1pt}
\begin{subfigure}[t]{0.24\textwidth}
        \centering
    \includegraphics[width=\textwidth]{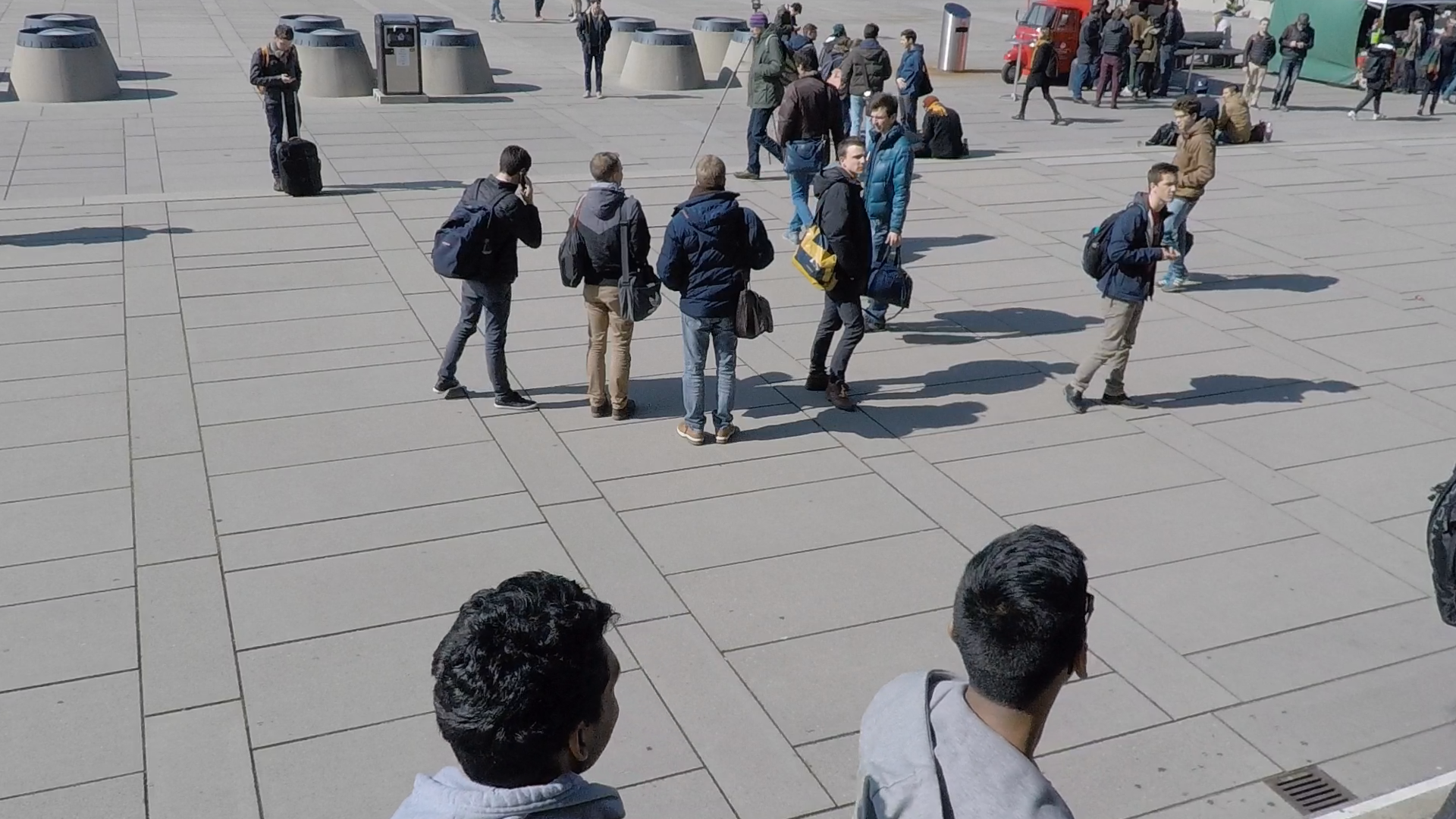}
    \caption{Camera 7.}
\end{subfigure}

\caption[A number of pictures.]{Synchronized corresponding frames from $7$ static cameras of the `WILDTRACK' dataset \cite{wild}. Four \emph{GoPro Hero $3$} and three \emph{GoPro Hero $4$} cameras are used for capturing the frames shown in the first and second row, respectively \cite{wild}. As seen from the figure, the data recording took place under sunny weather conditions and the height of camera positions is above average human height. The fields of vision of the cameras are also noticed to be overlapping, as discussed in Section \ref{sec:intro} (e.g. the concrete stairs in Camera 5 and 6). Occlusions between pedestrians are also observed (e.g. Camera 1 and 5) just as the `WILDTRACK' captures a densely crowded realistic scenario in front of ETH Zurich.} % The text in the square bracket is the caption for the list of figures while the text in the curly brackets is the figure caption
\label{fig:wild}
\end{figure*}

The `WILDTRACK' dataset provides annotations for $400$ synchronized frames for each of the $7$ static cameras, using a frame rate of $2$ fps. Note that original resolution of the images is $1920 \times 1080$ and on average, there are 20 pedestrians on each frame \cite{wild}. Due to unbalanced nature of labels in the `WILDTRACK' dataset, the loss function for the classification baseline and for the entire device-edge models is chosen to be weighted cross-entropy\footnote{We used the PyTorch implementation of \texttt{BCEWithLogitsLoss}, which is documented at: \url{https://pytorch.org/docs/stable/nn.html}}, which aims to optimize the metric:
\begin{equation}
    \mathrm{Balanced} \hspace{.2em}  \mathrm{Accuracy}=\frac{\mathrm{TPR}+\mathrm{TNR}}{2},
\end{equation}
for all class predictions, where $\mathrm{TPR}$ and $\mathrm{TNR}$ stand for True Positive Rate and True Negative Rate, respectively. \par

In order to ensure that one needs to effectively transmit the intermediate feature vectors coming from both cameras for configurations in Fig. \ref{fig:jdec} and \ref{fig:systems} , we define a correlation metric, $r_{\mathrm{corr}}$, to choose which pair of overlapping cameras to use for the evaluation of the proposed schemes\footnote{Clearly, we pick two distinct cameras to calculate this metric: $j \neq k$ in Equation \eqref{eq:corr}.}:
\begin{equation}
    r_{\mathrm{corr}}(j,k)=\frac{1}{N}\sum_{i=1}^{N}\frac{|\mathrm{Cam}_{j,i} \cup {\mathrm{Cam}_{k,i}}|}{
    |\mathrm{Cam}_{j,i} | + |\mathrm{Cam}_{k,i}| }\label{eq:corr}, 
\end{equation}
where $\mathrm{Cam}_{j,i}$ refers to the set of people appearing at Camera $j$ for $i^{th}$ frame. Note that $N=400$ for the `WILDTRACK' dataset. After calculating this correlation metric for all combinations of two cameras from the `WILDTRACK', we find that the pair of Camera $4$ and $5$ has the highest $r_{\mathrm{corr}}$ value of $0.911$\footnote{Since \begin{equation*}
    r_{\mathrm{corr}}(j,k) = \frac{1}{N}\sum_{i=1}^{N}\frac{ |\mathrm{Cam}_{j,i} | + |\mathrm{Cam}_{k,i}| - |\mathrm{Cam}_{j,i} \cap \mathrm{Cam}_{k,i} |}{
    |\mathrm{Cam}_{j,i} | + |\mathrm{Cam}_{k,i}| },
\end{equation*}
having a greater $r_{\mathrm{{corr}}}(j,k)$ indicates having, on average, less number of people simultaneously appearing at Camera $j$ and $k$.}. In order to make sure that the feature compression of the selected cameras is as nontrivial as possible, we choose Camera $4$ and $5$ to assess the performance of the proposed JSCC approaches for the task of person classification.

To make a fair comparison among the JSCC approaches discussed in Section \ref{sub:jscc}, we define an evaluation metric for the \emph{Single Users} scheme, $\mathrm{acc_{sep}}$, for a given bandwidth budget of $B$ as the following\footnote{Getting the weighted average as such is equivalent to computing the accuracy of the entire \emph{Single Users} scheme. This is useful because there is no need to train a new model for each combination of bandwidths; it suffices to train the individual networks for both cameras independently (see Fig. \ref{fig:sep}). This means we only need to train $2|B|$ times instead of $|B|^2$. }:

\begin{multline}
    \mathrm{acc_{sep}}(B) = \frac{1}{|E|} \sum_{{e \in E}} \max_{\substack{b_{1}, b_{2} \in \mathcal{B} \\ s.t. b_{1}+b_{2}=B}} (w_{4}\cdot \mathrm{acc}_{{4,b_{1}}}
    + w_{5}\cdot \mathrm{acc}_{{5,b_{2}}}),
\label{eq:best_comb}
\end{multline}

\vspace*{-10pt}

\begin{align*}
     \text{where } &w_{4} =\frac{|ID_{4}|}{|ID_{4}|+|ID_{5}|},  \\  &w_{5} =\frac{|ID_{5}|}{|ID_{4}|+|ID_{5}|}, \\
     &E: \text{set of independent experiments}, \\  &\mathcal{B }: \text{set of evaluated bandwidths}, \\
     &\mathrm{acc}_{k,b}: \text{mean balanced  accuracy across all } \\  &\text{classes for Camera $k$ } \text{at bandwidth } b, \\
     &ID_{k}: \text{set of all people appearing at Camera $k$}. 
\end{align*}
This metric ensures that the \emph{Single Users} has the flexibility to choose the best bandwidth allocation given a bandwidth budget of $B$ --- this provides an upper bound on the performance that can be achieved by the JSCC scheme shown in Fig. \ref{fig:sep}. We will refer to this metric as \emph{Optimal Combination} in the following parts. 

%\begin{tabular}{p{2.7cm} p{5.5cm}
\begin{table*}
  \begin{center}
  \begin{adjustbox}{width=.9\textwidth} %width=.75
    \begin{tabular}{p{2.7cm} p{11.2cm}} % <-- Alignments: 1st column left, 2nd middle and 3rd right, with vertical lines in between
    \hline
      \textbf{Short Name} & \textbf{JSCC Scheme}   \\
      \hline
       Single Users + Eq.BW &  \emph{Single Users} with equal bandwidth allocation for both users \\
      Single Users + Opt.BW & \emph{Single Users} evaluated with 
      \emph{Optimal Combination} \\
      J-Dec + OMA & \emph{Joint Decoding + Orthogonal Multiple Access}
      with equal power allocation for both users \\
      J-Dec + NOMA &  \emph{Joint Decoding + Non-Orthogonal Multiple Access} 
      with  equal power allocation for both users \\
      J-Enc1 &  \emph{Joint Encoding} \\
      J-Enc2 &  \emph{Joint Encoding + Non-Orthogonal Transmission}\\
      
      \hline
    \end{tabular}
    \end{adjustbox}
  \end{center}
  \caption{Table of definition for the evaluated JSCC approaches.}
  \label{table_def}
\end{table*}

We provide a table of definitions for the evaluated JSCC approaches in Table \ref{table_def}. In our experiments, we use the same SNR value for the training and testing phases. In the following parts, the results shown for the JSCC approaches are the median (i.e. Q2) of 100 different evaluations carried out for each bandwidth and SNR combination considered. If shown, the error bar edges correspond to the lower and upper quartiles (i.e. Q1 and Q3). For the digital scheme, the best accuracy is kept across different values of $C$ in Equation \eqref{eq:shannon}.

%\hl{the average of 100 different evaluations} 

%eskiden tek column ve [H] vardi

\subsection{Performance for Different Methods}

\begin{figure}[t]
    \centering
    \begin{subfigure}[t]{0.49\textwidth}
        \centering
    \includegraphics[width=\textwidth]{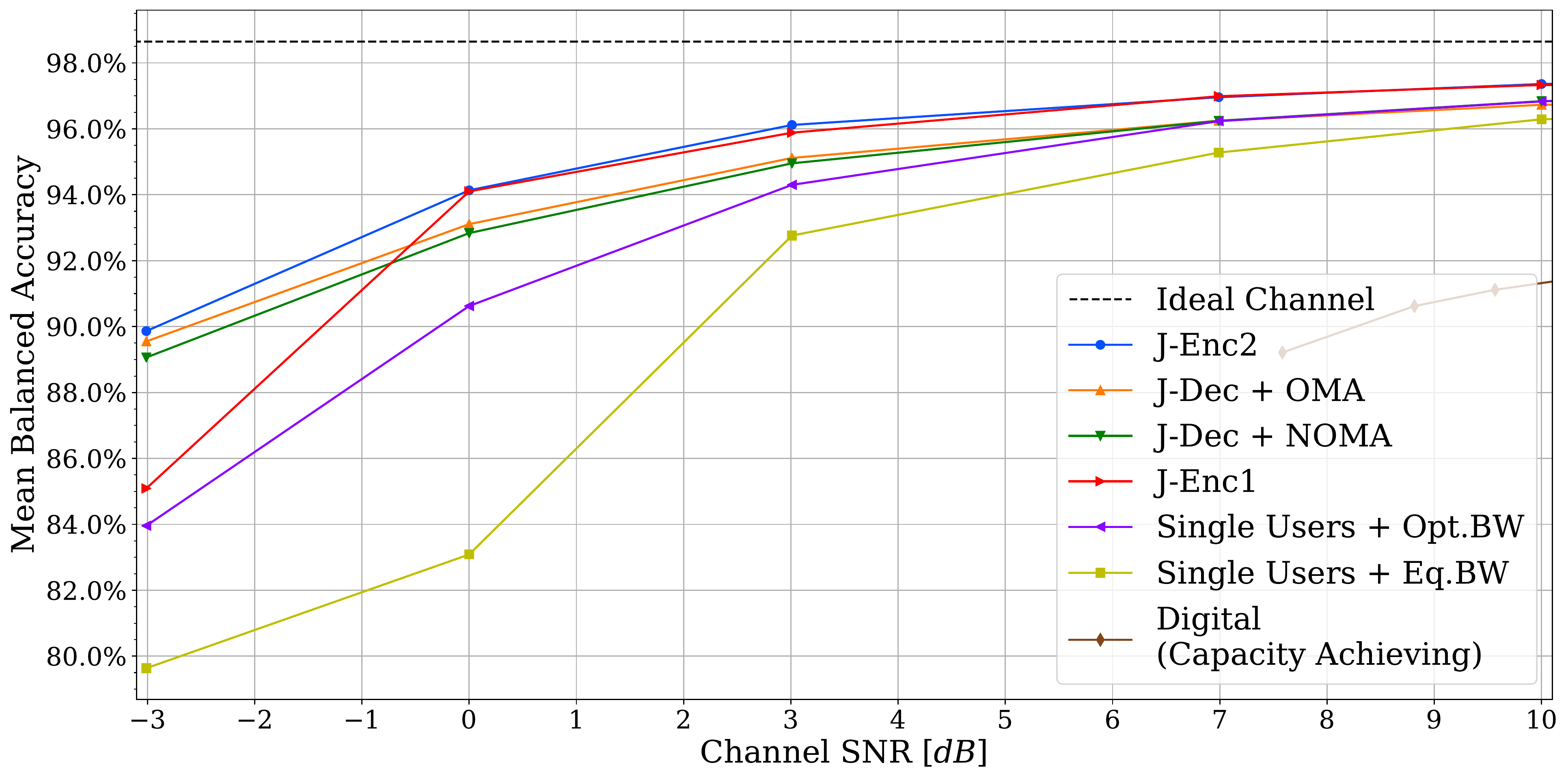}
    \caption{$B=32$}
    \label{fig:dif_method1}
    \end{subfigure}
    \begin{subfigure}[t]{0.49\textwidth}
        \centering
    \includegraphics[width=\textwidth]{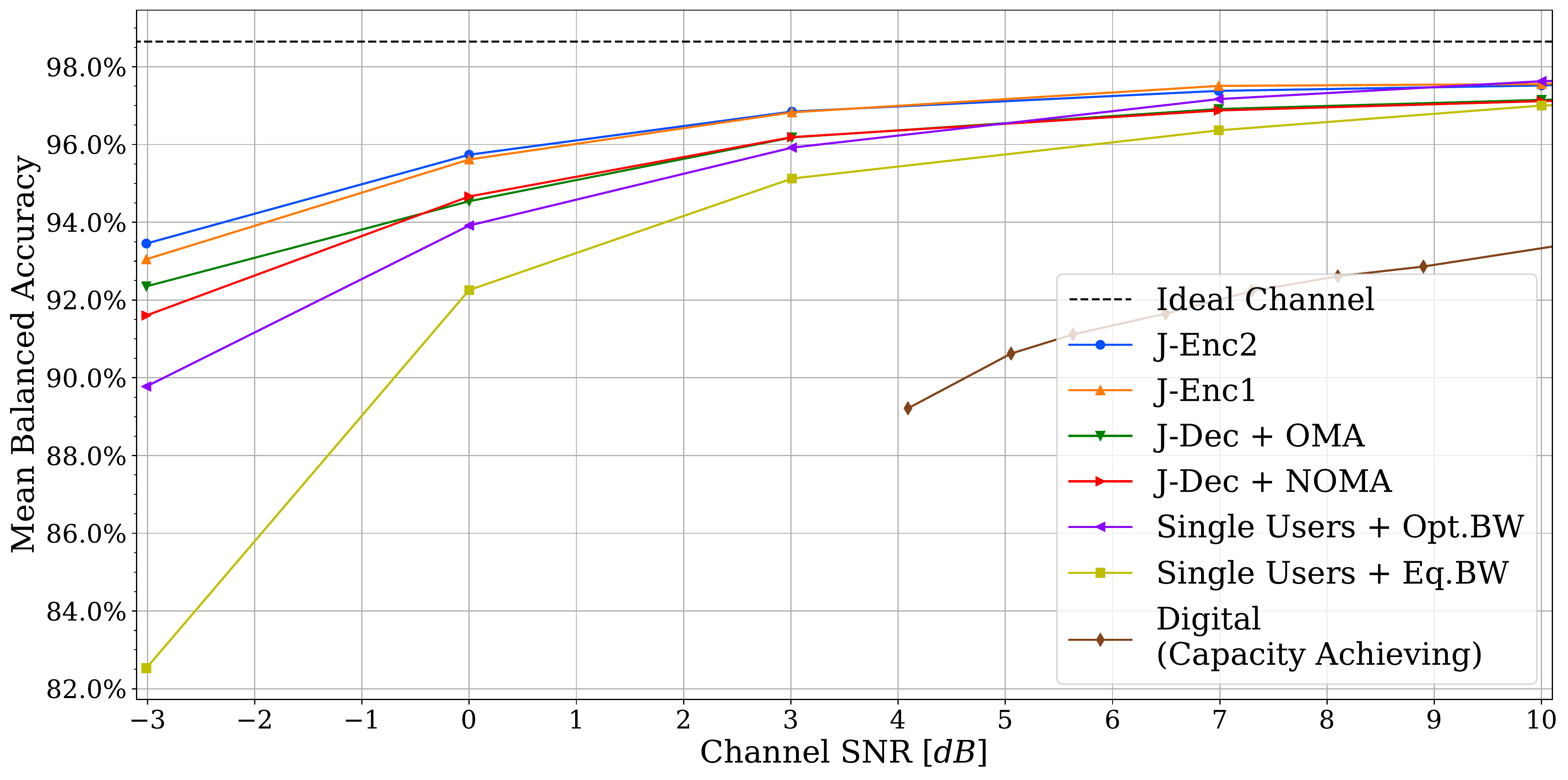}
    \caption{$B=48$}
    \label{fig:dif_method2}
    \end{subfigure}
    \begin{subfigure}[t]{0.49\textwidth}
        \centering
    \includegraphics[width=\textwidth]{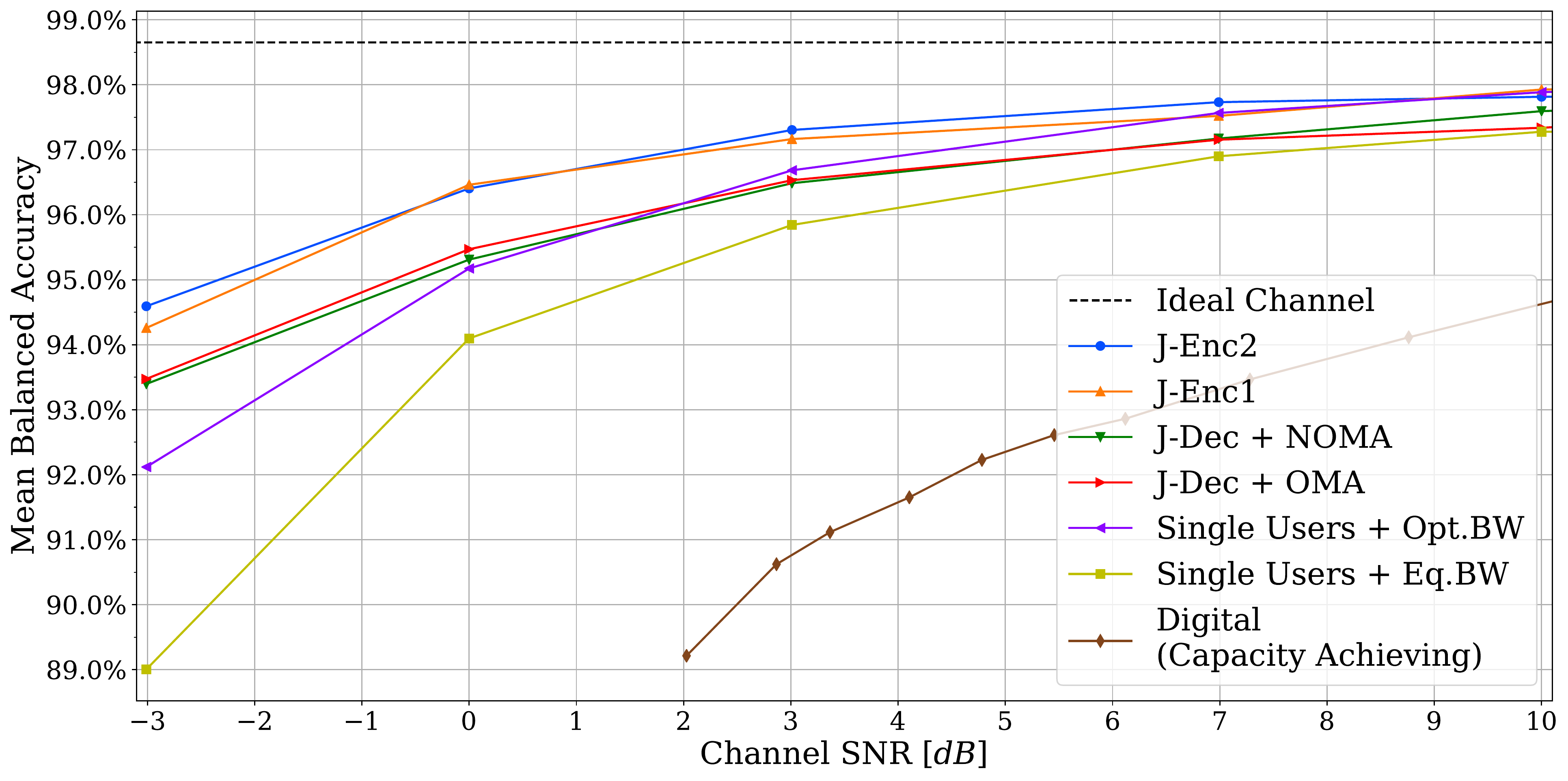}
    \caption{$B=64$}
    \label{fig:dif_method3}
    \end{subfigure}
    \caption{Comparison of performance for different approaches for $B \in \{32,48,64\}$. For the \emph{Single Users + Opt.BW} scheme, the metric \emph{Optimal Combination}
    is evaluated for $\mathcal{B} \in \{ 2, 4, 6, \ldots , 62\}$ in Equation \eqref{eq:best_comb}. }
    \label{fig:dif_method}
\end{figure}

We plot the accuracy achieved by the JSCC and digital schemes as a function of various channel SNRs in  Fig. \ref{fig:dif_method1}, \ref{fig:dif_method2} and \ref{fig:dif_method3}. As seen, the JSCC approaches outperform both the digital scheme and Single Users + Eq.BW in all of the scenarios considered.  Differently from the digital scheme, the J-Enc1, J-Dec + OMA and J-Dec + NOMA closely approach J-Enc2 at the high SNR regime, but fail to achieve a similar performance for smaller values of SNR. For bandwidth budgets of  $B \in \{32,48,64\}$, the schemes J-Enc2, J-Enc1, J-Dec + OMA and J-Dec + NOMA perform better than the Single Users + Opt.BW, notably for smaller values of SNR.

\subsection{Performance for Different Channel SNRs}

\begin{figure*}
    \centering
    \begin{subfigure}{0.49\textwidth} 
        \centering
    \includegraphics[width=\textwidth]{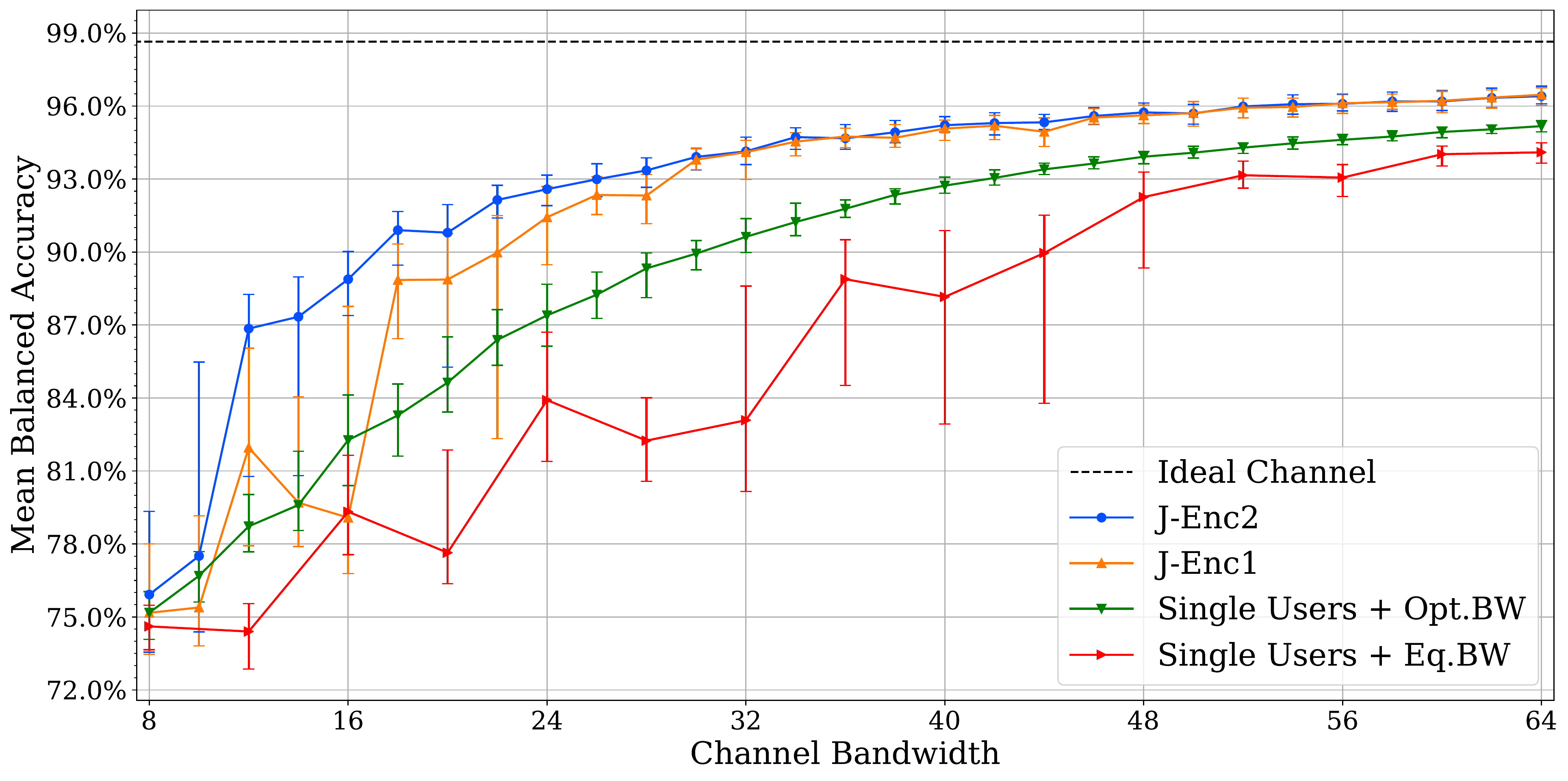}
    \caption{$\mathrm{{SNR}_{train}}=\mathrm{{SNR}_{test}}= 0$   $\mathrm{dB} $}
    \label{fig:snr0_1}
    \end{subfigure}
    \begin{subfigure}{0.49\textwidth}
        \centering
    \includegraphics[width=\textwidth]{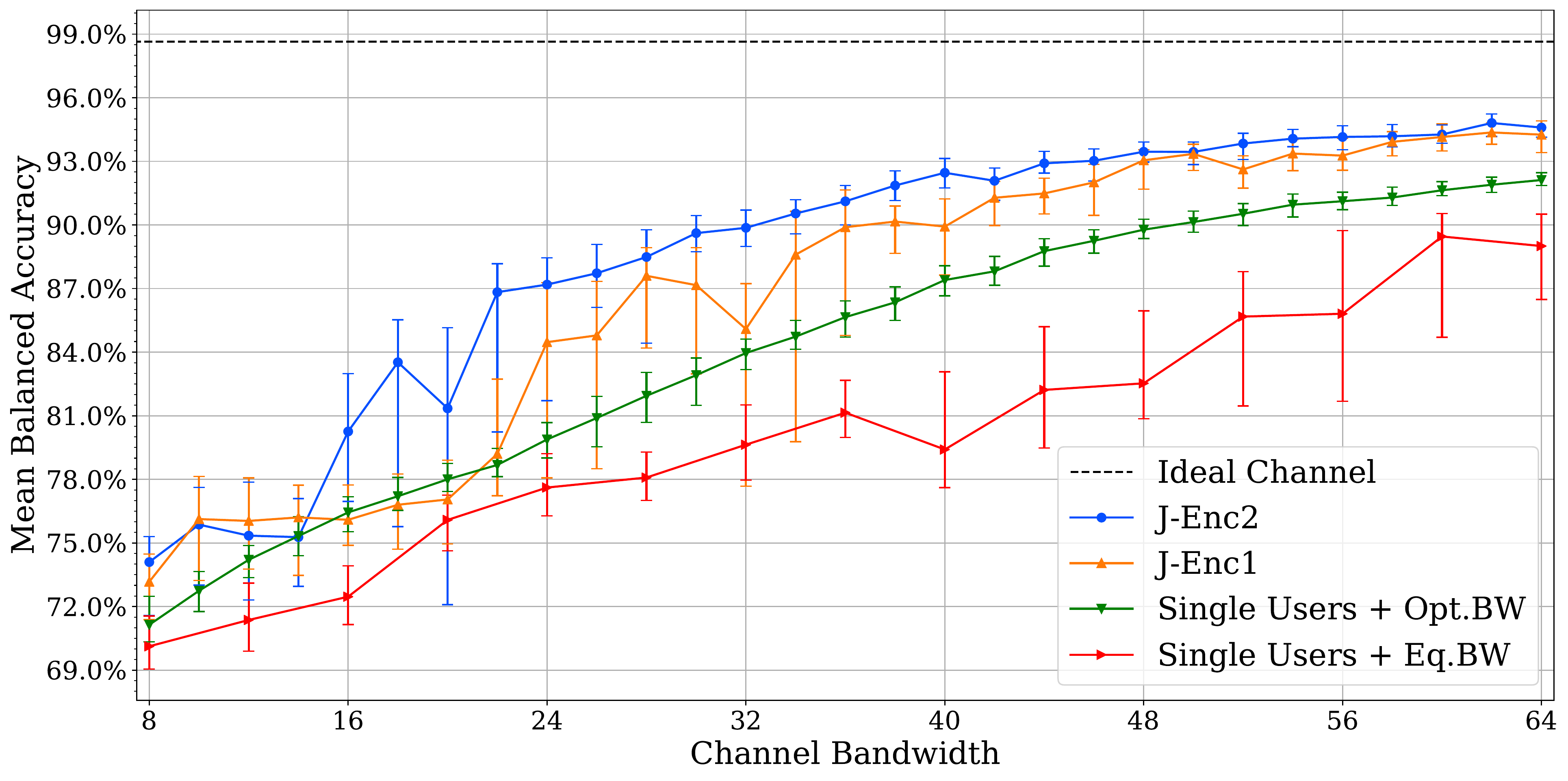}
   \caption{$\mathrm{{SNR}_{train}}=\mathrm{{SNR}_{test}}= -3$   $\mathrm{dB} $}
    \label{fig:snr-3_1}
    \end{subfigure}
    \caption{Accuracy as a function of the channel bandwidth for $\mathrm{SNR} \in \{-3,0\}$ $\mathrm{dB}$.
    For the \emph{Single Users + Opt.BW} scheme, the metric \emph{Optimal Combination}
    is evaluated for $\mathcal{B} \in \{ 2, 4, 6, \ldots , 62\}$ in Equation \eqref{eq:best_comb}.}
\label{fig:SNRs_1}
\end{figure*}

\begin{figure*}[ht!]
    \centering
    \begin{subfigure}{0.49\textwidth} 
        \centering
    \includegraphics[width=\textwidth]{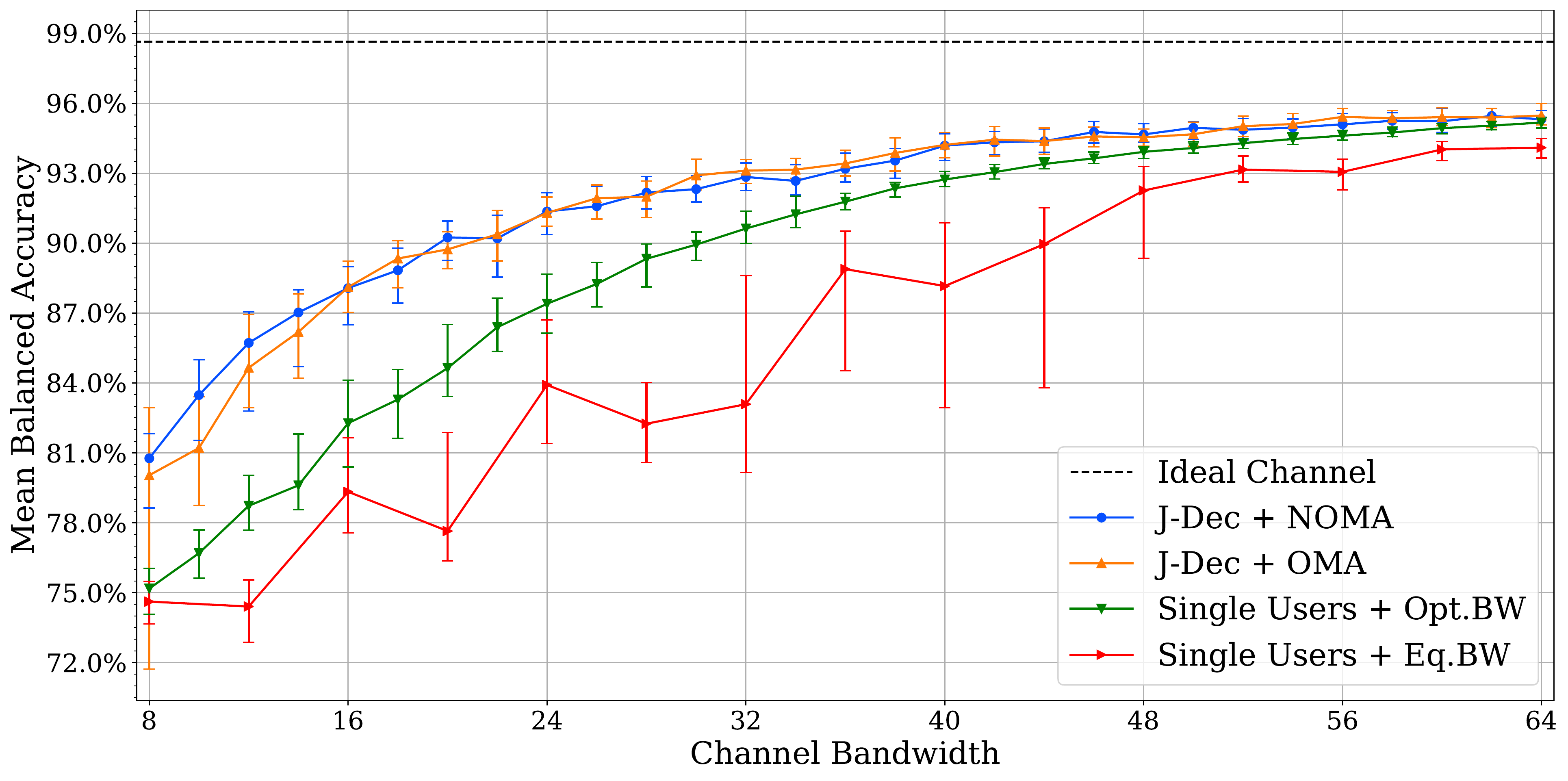}
    \caption{$\mathrm{{SNR}_{train}}=\mathrm{{SNR}_{test}}= 0$   $\mathrm{dB} $}
    \label{fig:snr0_2}
    \end{subfigure}
    \begin{subfigure}{0.49\textwidth}
        \centering
    \includegraphics[width=\textwidth]{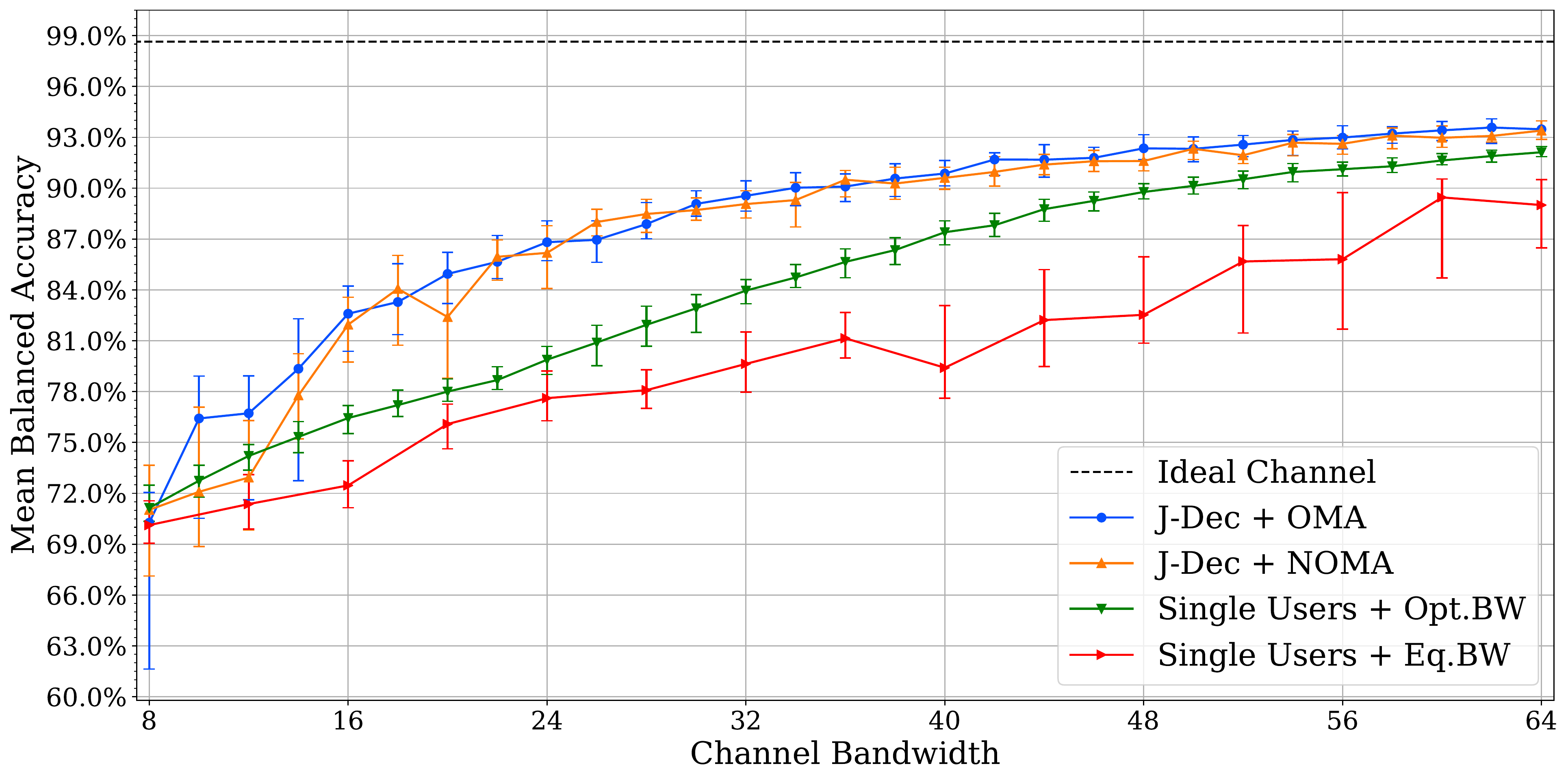}
   \caption{$\mathrm{{SNR}_{train}}=\mathrm{{SNR}_{test}}= -3$   $\mathrm{dB} $}
    \label{fig:snr-3_2}
    \end{subfigure}
    \caption{Accuracy as a function of the channel bandwidth for $\mathrm{SNR} \in \{-3,0\}$ $\mathrm{dB}$.
    For the \emph{Single Users + Opt.BW} scheme, the metric \emph{Optimal Combination}
    is evaluated for $\mathcal{B} \in \{ 2, 4, 6, \ldots , 62\}$ in Equation \eqref{eq:best_comb}.}
\label{fig:SNRs_2}
\end{figure*}

\begin{figure*}[ht!]
    \centering
    \begin{subfigure}{0.49\textwidth} 
        \centering
    \includegraphics[width=\textwidth]{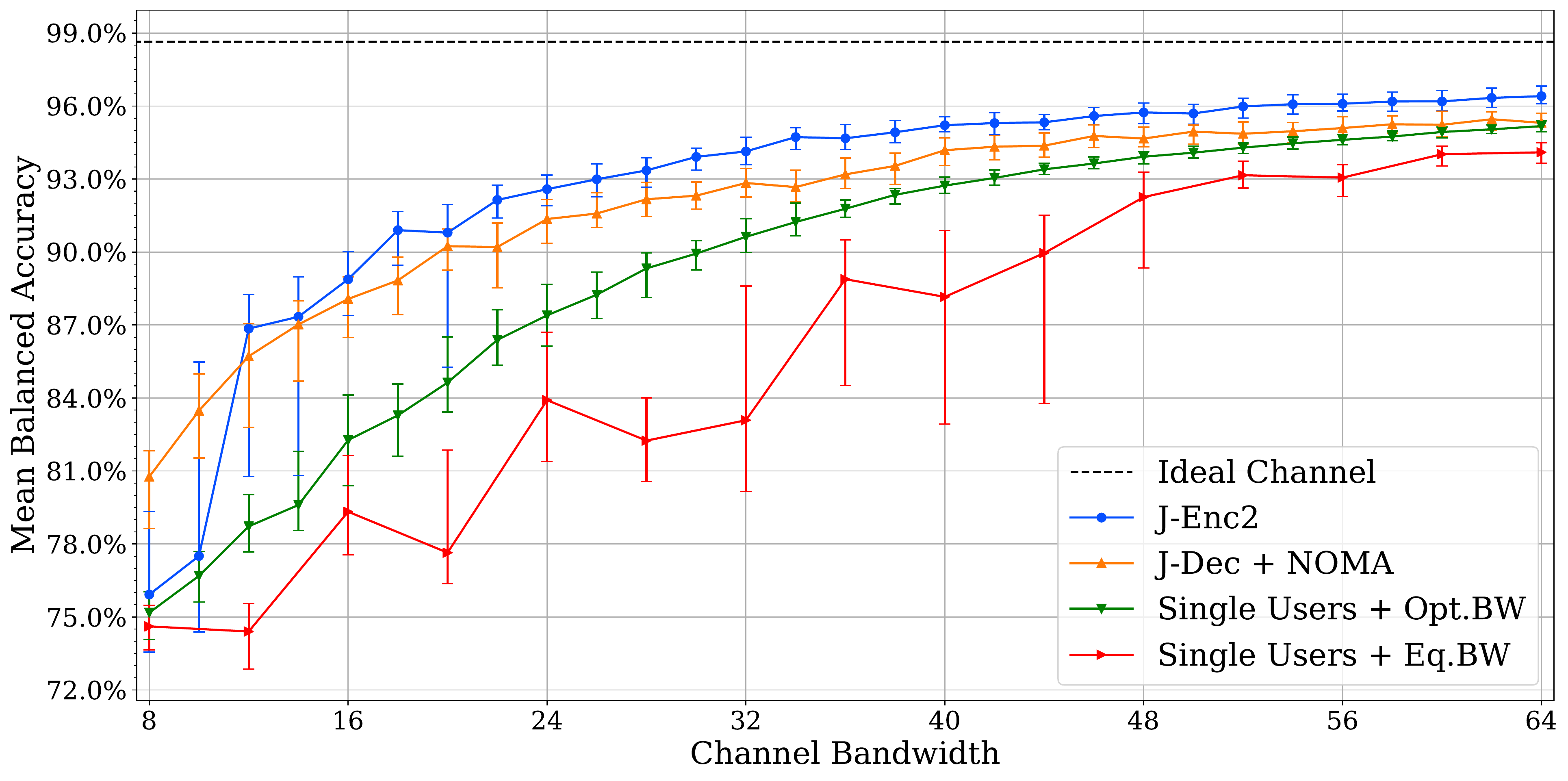}
    \caption{$\mathrm{{SNR}_{train}}=\mathrm{{SNR}_{test}}= 0$   $\mathrm{dB} $}
    \label{fig:snr0_3}
    \end{subfigure}
    \begin{subfigure}{0.49\textwidth}
        \centering
    \includegraphics[width=\textwidth]{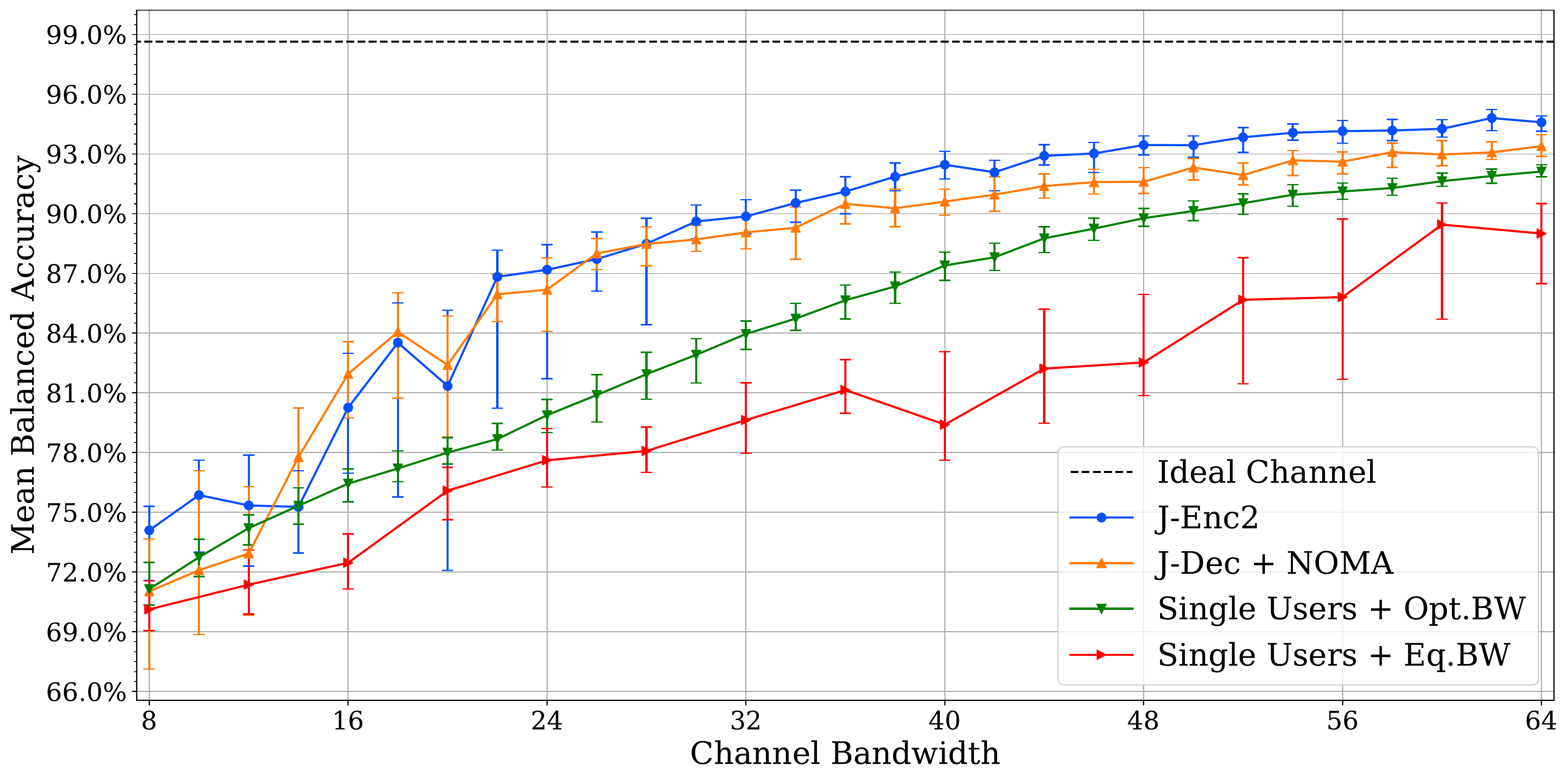}
   \caption{$\mathrm{{SNR}_{train}}=\mathrm{{SNR}_{test}}= -3$   $\mathrm{dB} $}
    \label{fig:snr-3_3}
    \end{subfigure}
    \caption{Accuracy as a function of the channel bandwidth for $\mathrm{SNR} \in \{-3,0\}$ $\mathrm{dB}$.
    For the \emph{Single Users + Opt.BW} scheme, the metric \emph{Optimal Combination}
    is evaluated for $\mathcal{B} \in \{ 2, 4, 6, \ldots , 62\}$ in Equation \eqref{eq:best_comb}.}
\label{fig:SNRs_3}
\end{figure*}

In this experiment, we investigate the effect of varying the channel SNR on the JSCC approaches. The accuracy as a function of channel bandwidth is shown in Fig. \ref{fig:SNRs_1}, \ref{fig:SNRs_2} and \ref{fig:SNRs_3} for diﬀerent SNR values of $0$ and $-3$ dB. As seen in Fig. \ref{fig:SNRs_1} and \ref{fig:SNRs_3}, the J-Enc2 achieves the best performance across the bandwidth range considered for both SNR values. Although the Single Users + Opt.BW closely approaches the J-Dec + NOMA and J-Enc2 with increasing channel bandwidth allocation, it fails to be as robust as the other proposed JSCC schemes, especially for $\mathrm{SNR}=-3$ $\mathrm{dB}$. From Fig. \ref{fig:SNRs_2}, we can conclude that the schemes J-Dec + OMA and J-Dec + NOMA perform similarly across evaluated channel bandwidth values, both achieving better accuracy than the Single Users + Opt.BW. As seen in Fig. \ref{fig:SNRs_3}, one can argue that although the J-Enc2 scheme beats the J-Dec + NOMA for evaluated channel bandwidth and SNR ranges, the reason that the schemes J-Enc2 and J-Dec + NOMA perform better than the Single Users + Opt.BW is due to joint decoding.

\subsection{Performance for Different Bandwidths}

\begin{figure*}[ht]
        \centering
        \includegraphics[width=0.49\textwidth]{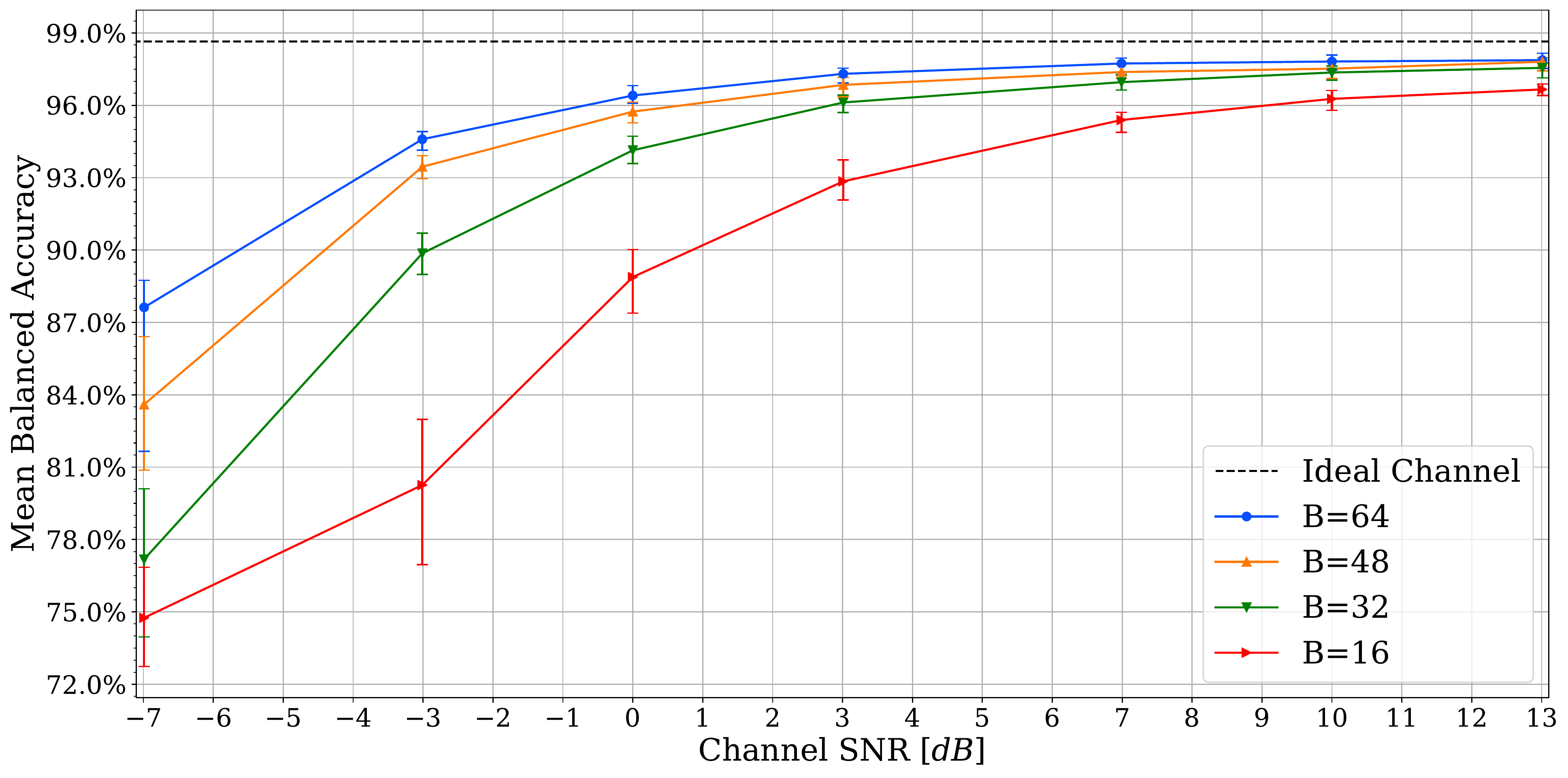}
    \caption{Accuracy as a function of the channel SNR for $B \in \{16, 32, 48, 64\}$. The more bandwidth is allocated for the JSCC, the more robust it becomes against the channel noise.}
    \label{fig:bandwidth}
\end{figure*}

In the last experiment, we examine the effect of the channel bandwidth on the J-Enc2. The accuracy as a function of channel SNR is plotted in Fig. \ref{fig:bandwidth} for diﬀerent channel bandwidth values of 16, 32, 48, 64. It can be observed that the accuracy increases considerably as more channel bandwidth is allocated, but the relative gain diminishes as the channel bandwidth approaches the original feature dimension. From Fig. \ref{fig:bandwidth}, it can be claimed that the bandwidth choice  of $B=48$ is best in terms of balancing the trade-off between the accuracy and allocated bandwidth.

%It can be seen that the accuracy and robustness increases signiﬁcantly with the bandwidth, but the relative gain becomes smaller as we approach the original feature vector dimension. Therefore, from accuracy-bandwidth trade-oﬀ perspective the best choice is to aim for a bandwidth of B = 512 or B = 256 as they provide a signiﬁcant accuracy gain, while still operating over a reasonable bandwidth

\section{Conclusions}

In this work, we studied person classification task at the wireless edge carried out by power-constrained multi-view cameras, with overlapping fields of view. Firstly, we introduced a digital (separate) scheme which employed quantization and arithmetic coding for feature compression. Aiming to exploit the correlation between the fields of view of the cameras by joint encoding and/or joint decoding, we proposed JSCC schemes for robust transmission of feature vectors. These schemes incorporated an autoencoder-based architecture for intermediate feature compression. The JSCC approaches achieve a satisfactory classification accuracy even with extremely limited bandwidth and power constraints. This was achieved with a multi-step training strategy. The JSCC schemes also achieve superior results in comparison to the digital scheme under the evaluated SNR and bandwidth constraints. As a future work, we plan to provide an extensive evaluation of the proposed approaches at various DNN splitting points as well as to investigate various layers and different activation functions for the autoencoder architecture. In order to address the limited computational resources of the power-constrained IoT devices, we will also aim to incorporate pruning into the joint training phase.

\bibliographystyle{ieeetr}
%\bibliography{IEEEabrv,refs}
%\clearpage
\bibliography{refs}

\begin{thebibliography}{10}

\bibitem{mck}
A.~R. Fredrik~Dahlqvist, Mark~Patel and J.~Shulman, {\em Growing opportunities
  in the Internet of Things}, 2019 (Accessed June 4, 2020).
\newblock
  \url{https://www.mckinsey.com/industries/private-equity-and-principal-investors/our-insights/growing-opportunities-in-the-internet-of-things#:~:text=The\%20number\%20of\%20businesses\%20that,almost\%20threefold\%20increase\%20from\%202018.&text=1.}

\bibitem{edge:2}
A.~{Erfan Eshratifar}, A.~{Esmaili}, and M.~{Pedram}, ``{BottleNet: A Deep
  Learning Architecture for Intelligent Mobile Cloud Computing Services},''
  {\em arXiv e-prints}, p.~arXiv:1902.01000, Feb. 2019.

\bibitem{edge:3}
J.~{Shao} and J.~{Zhang}, ``{BottleNet++: An End-to-End Approach for Feature
  Compression in Device-Edge Co-Inference Systems},'' {\em arXiv e-prints},
  p.~arXiv:1910.14315, Oct. 2019.

\bibitem{mj:1}
M.~{Jankowski}, D.~{Gunduz}, and K.~{Mikolajczyk}, ``{Deep Joint
  Transmission-Recognition for Power-Constrained IoT Devices},'' {\em arXiv
  e-prints}, p.~arXiv:2003.02027, Mar. 2020.

\bibitem{jscc:1}
E.~{Bourtsoulatze}, D.~{Burth Kurka}, and D.~{Gunduz}, ``{Deep Joint
  Source-Channel Coding for Wireless Image Transmission},'' {\em arXiv
  e-prints}, p.~arXiv:1809.01733, Sept. 2018.

\bibitem{resnet}
K.~He, X.~Zhang, S.~Ren, and J.~Sun, ``Deep residual learning for image
  recognition,'' in {\em Proceedings of the IEEE Conf. on computer vision and
  pattern recognition}, pp.~770--778, 2016.

\bibitem{quant}
R.~M. {Gray} and D.~L. {Neuhoff}, ``Quantization,'' {\em IEEE Transactions on
  Information Theory}, vol.~44, no.~6, pp.~2325--2383, 1998.

\bibitem{mj:2}
M.~{Jankowski}, D.~{Gunduz}, and K.~{Mikolajczyk}, ``{Deep Joint Source-Channel
  Coding for Wireless Image Retrieval},'' {\em arXiv e-prints},
  p.~arXiv:1910.12703, Oct. 2019.

\bibitem{jscc:2}
J.~{Ball{\'e}}, V.~{Laparra}, and E.~P. {Simoncelli}, ``{Density Modeling of
  Images using a Generalized Normalization Transformation},'' {\em arXiv
  e-prints}, p.~arXiv:1511.06281, Nov. 2015.

\bibitem{jscc:4}
J.~{Ball{\'e}}, V.~{Laparra}, and E.~P. {Simoncelli}, ``{End-to-end Optimized
  Image Compression},'' {\em arXiv e-prints}, p.~arXiv:1611.01704, Nov. 2016.

\bibitem{jscc:3}
J.~{Ball{\'e}}, D.~{Minnen}, S.~{Singh}, S.~J. {Hwang}, and N.~{Johnston},
  ``{Variational image compression with a scale hyperprior},'' {\em arXiv
  e-prints}, p.~arXiv:1802.01436, Jan. 2018.

\bibitem{jankowski2020wireless}
M.~Jankowski, D.~Gunduz, and K.~Mikolajczyk, ``Wireless image retrieval at the
  edge,'' 2020.

\bibitem{wild}
T.~{Chavdarova}, P.~{Baqué}, S.~{Bouquet}, A.~{Maksai}, C.~{Jose},
  T.~{Bagautdinov}, L.~{Lettry}, P.~{Fua}, L.~{Van Gool}, and F.~{Fleuret},
  ``Wildtrack: A multi-camera hd dataset for dense unscripted pedestrian
  detection,'' in {\em 2018 IEEE/CVF Conference on Computer Vision and Pattern
  Recognition}, pp.~5030--5039, 2018.

\bibitem{PETS}
J.~{Ferryman} and A.~{Shahrokni}, ``Pets2009: Dataset and challenge,'' in {\em
  2009 Twelfth IEEE International Workshop on Performance Evaluation of
  Tracking and Surveillance}, pp.~1--6, 2009.

\bibitem{epfl}
T.~{Chavdarova} and F.~{Fleuret}, ``{Deep Multi-camera People Detection},''
  {\em arXiv e-prints}, p.~arXiv:1702.04593, Feb. 2017.

\end{thebibliography}

\end{document}